\documentclass[11pt]{article}
\usepackage{graphicx}
\usepackage{amssymb,amsmath}
\usepackage{fullpage}
\usepackage[small]{caption}
\usepackage{hyperref}
\usepackage{fancyhdr}
\usepackage[natbibapa]{apacite}
\usepackage{setspace}
\def\mask{\underline{\ \ \ \ \ }\ }

\title{\textbf{Word meaning in minds and machines}}
\date{}
\author{
	\textbf{Brenden M. Lake}\\
	New York University\\
	Facebook AI Research
	\and
	\textbf{Gregory L. Murphy}\\
	New York University\\
}

\fancypagestyle{alim}{\fancyhf{}\fancyhead[L]{In press at \emph{Psychological Review}}}

\begin{document}
\maketitle

\thispagestyle{alim}

\begin{abstract}
\noindent Machines have achieved a broad and growing set of linguistic competencies, thanks to recent progress in Natural Language Processing (NLP). Psychologists have shown increasing interest in such models, comparing their output to psychological judgments such as similarity, association, priming, and comprehension, raising the question of whether the models could serve as psychological theories. In this article, we compare how humans and machines represent the meaning of words. We argue that contemporary NLP systems are fairly successful models of human word similarity, but they fall short in many other respects. Current models are too strongly linked to the text-based patterns in large corpora, and too weakly linked to the desires, goals, and beliefs that people express through words. Word meanings must also be grounded in perception and action and be capable of flexible combinations in ways that current systems are not. We discuss more promising approaches to grounding NLP systems and argue that they will be more successful with a more human-like, conceptual basis for word meaning.
\end{abstract}

\section{Introduction}

Psychological semantics is the study of how people represent the meanings of words and then build sentence meaning out of those representations. People use language dozens of time a day---to have conversations and give instructions, to read and write, to label objects and teach. A theory of psychological semantics must provide the basis for how people do all those things, choosing which words to use and understanding the words they read or hear. In this article we focus on the mental representation of word meaning.

Human language is still the gold standard for a communication system, but artificial intelligence (AI) systems have made important progress in language use. Research on \emph{Natural Language Processing} (NLP) develops systems that understand language to the degree that computers can carry out useful tasks. As described below, such systems use vast text corpora to learn about words, using neural networks and other statistical models. The recent explosion of research in NLP, driven largely by advances in neural networks (also called \emph{deep learning}), has resulted in constantly improving performance on various benchmarks that require interpreting words and sentences. Systems are now used in interfaces with customers to make sales or solve problems. Some systems even perform tasks that were historically assumed to be solely within the purview of humans, such as translation, summarization, question answering, and natural language inference. 

One way to think about such progress is merely in terms of engineering: There is a job to be done, and if the system does it well enough, it is successful. Engineering is important, and it can result in better and faster performance and relieve humans of dull labor such as keying in data or making airline itineraries or buying socks. Ultimately, tasks such as machine translation, automatic summarization, and human-machine communication may change our world for the better. Doing these things can certainly be described as semantic processing, but they may not be the same semantic processing that human speakers and listeners engage in. 

The enormous progress of these systems has led some researchers to suggest that they are potential models of psychological semantics (Section \ref{nlp_as_theory}). That is, the representations that they derive for words are functionally similar to those that people derive through language learning. A stronger (and more implausible) claim is that the way people \emph{learn} word meanings is similar to the way models do. We will focus on the first claim here, which is perhaps surprisingly held by a number of psychologists, or at least considered as a viable hypothesis. Many AI researchers do not dwell on whether their models are human-like; few would complain that a highly accurate machine translation system doesn't do things the way human translators do. Nonetheless, progress in NLP is usually evaluated, at least implicitly, against some human standard, raising the real possibility that progress in NLP translates to progress on modeling how people understand the meaning of words. We will argue that contemporary NLP techniques may indeed do many things well, but models will need to push beyond current popular approaches in order to provide a theory of psychological semantics. 

We will not suggest that NLP should redirect its efforts to building models of psychological semantics. As we discuss, NLP technologies have been tremendously successful at many language tasks without worrying about the psychological plausibility of their semantic representations. For many applications, this engineering-driven approach will be sufficient. In other cases, we see strong potential for improving NLP systems by taking a more psychological approach to word meaning. That is, although NLP models keep getting more successful, there may be limits on their performance in comprehending and producing language that could be overcome with representations that are more like those people have. And if these models are intended as psychological theories, they will need to go beyond current benchmarks to considerations of how people use language every day, as we will argue below.

\subsection{Coverage of this article}
We begin by briefly reviewing approaches to word meaning in the psychological literature (Section \ref{sec_semantics_cogsci}) and introducing desiderata for models of psychological semantics (Section \ref{sec_intro_desiderata}). We then cover NLP approaches to learning word representations, starting with those initially constructed by psychologists and then moving to more recent NLP models in Machine Learning that learn both word and sentence representations  (Section \ref{sec_NLP_models}).
Our main question is whether NLP systems are likely to be successful in representing human semantic knowledge. We argue that current means of representing words are useful for modeling word similarity, although the details don't always align with human semantic similarity (Section \ref{sec_similarity}). These word representations, however, are not adequate to support the flexible behaviors for which people rely on their semantic representations. We discuss five such classes of human behavior and the challenges they present for models of word meaning (Section \ref{sec_main_desiderata}). We end by discussing the possibility that building more realistic models of psychological semantics will also improve NLP's ability to handle more advanced and difficult problems of understanding and communication (Section \ref{sec_conclusion}).

\section{Semantics in Cognitive Science} \label{sec_semantics_cogsci}
Before discussing word meaning in NLP, we first take a brief tour of the approaches to semantics in linguistics and psychology in order to understand what it is that such theories must explain. Within most of philosophy and linguistics, semantics is \emph{referential}. That is, linguistic meaning is analyzed as a relationship between words and the world, and sentence meaning describes a state of affairs that can be mapped to situations in the world \citep[e.g.,][]{ChierchiaGennaroandMcConnell-Ginet1990}. For example, a word like \emph{dog} has a meaning that allows you to pick out all the dogs in the world. To a first approximation, the meaning is the set of all such dogs, and if you use the word to refer to a member of that set, you are using it correctly (and literally). People who fully understand the meaning of \emph{dog} would name all and only dogs with the word (excepting uninteresting cases such as not fully seeing the object, being incapacitated in some way, etc.). A problem with this view, however, is that dogs are coming into and going out of existence at a rapid rate. Many thousands of dogs are born and die every day. Thus, the set of dogs is constantly changing from moment to moment. That is not a very stable basis for a word meaning, which doesn't intuitively seem to be changing at all. Indeed, an implication of a simple referential theory would be that the meaning of \emph{dog} is today completely disjoint from what it was 30 years ago, assuming that no member of that set of dogs is still with us today. That does not seem correct. 

For this reason, formal linguists have developed more complicated analyses of meaning, such as claiming that \emph{dog} picks out sets of objects in an infinite number of \emph{possible worlds} \citep[see ][]{Dowty1981}. These worlds are keyed to time and context, such that the extension of \emph{dog} varies depending on the circumstances. Such a conception also allows us to refer to dogs within possible worlds that do not actually exist, such as hypothetical ones (``If there were no cats, dogs would still find something to chase.'') or fictional ones. Hypothetical and fictional situations are simply more possible worlds, which speakers may refer to. 

Why do (some) linguists insist on these kinds of analyses, which often result in seemingly unilluminating statements such as ``The meaning of \emph{dog} is the denotation [[dog]]''? The reason is an extremely powerful one, namely that the utility of language is in its ability to provide information about actual things in the world---to draw our attention to those things, learn about them, and then to take action on them \citep[see][Ch. 1]{ChierchiaGennaroandMcConnell-Ginet1990}. Language is not a parlor game in which we only utter formulas of statement and response. Language presumably evolved because of situations in which people can say things like, ``Look for blueberries on the other side of the hill,'' or ``Watch out for that car!'' or ``I love you and want to spend my life with you.'' The significance of such statements lies in their ability to communicate life-saving and life-improving information. Information must relate to the world if it is to be helpful. Talking about blueberries is only useful if doing so actually directs us to a particular kind of edible fruit; warning about a car is only helpful if the hearer then looks out for a car, rather than for blueberries. If word meanings did not relate to our world, they would not be helpful. 

The fact that language refers to the world seems indisputable, but exactly how to capture that relation is not as clear. For psychologists, the problem with the referential approach to meaning is that the possible-world semantics cannot be something that humans do in their heads. Even if a speaker is accurate in her use of the word \emph{dog}, she cannot keep the entire set of the current world's dogs in her head, much less the sets of dogs in all of the infinite number of possible worlds. Philosophers since Frege have talked about another aspect of meaning besides the denotative aspect: \emph{senses} (sometimes referred to as \emph{intensions}). The sense of a word is its ``mode of presentation,'' as \citeauthor{Frege1892} (1892/1960) referred to it, which can be understood as a way of thinking about the word. For many psychologists, this has been interpreted as a kind of mental description. A speaker cannot keep a representation of every dog in the world in her head, but she can keep a description of what dogs are like, which then enables her to apply the word to them. We know that dogs have four legs, teeth, and fur, weigh 5 to 75 pounds, like to chase smaller animals and cars, bark, eat scraps that fall on the floor, and so on. We have detailed knowledge of the faces, colors, proportions, and sounds of dogs, along with memories of individual dogs. Although one might argue that some of this knowledge is not strictly part of the meaning of \emph{dog} (e.g., that dogs are likely to snap up food that falls on the floor or personal memories), such knowledge allows one to pick out many (though likely not all) of the dogs in the world under variable circumstances. Thus, this mental description of the word allows people to refer to things in the world and to communicate with other people who have similar such descriptions associated to the word.

Language directly connects to our knowledge of the world, as attempts to make realistic language processing systems discovered early on \citep{Schank1977}. For example, imagine the following conversation. 
\begin{quote}
Marjorie: I can't come to the reception after the talk, because of Fred.

Todd: Fred?

Marjorie: Fred is my dog.
\end{quote}
The mere statement that Fred is Marjorie's dog explains many things. We realize that dogs require frequent maintenance. They must be fed regularly and usually must be let outside multiple times a day. Furthermore, dogs are social creatures and do not take well to being left alone for very long stretches of time. None of this is actually said in that conversation, yet Marjorie expects that Todd will understand at least some of these things and therefore infer that she can't go to the reception because of her need to let Fred out, feed him, and so on. That is, merely by saying the word \emph{dog}, Marjorie allows Todd to access his knowledge of the dogs and their properties, which he can then use to make necessary inferences to understand her explanation. If language did not have this property, then conversations would be inordinately long and laborious.

This example raises the question of where to draw the line between semantics of language and world knowledge. Different disciplines have come to very different answers to this question, ranging from a very minimal account of word meaning (e.g., \emph{dog} means the property of being a dog; \citeauthor{Fodor1998}, \citeyear{Fodor1998}) to accounts in which there is no demarcation between word meaning and general knowledge \citep{Elman2014}. There was a time when it could be suggested that a word meaning was simply a definition that determined when the word could be used. However, with the wide belief that definitions do not exist for many words \citep{Rosch1978,Wittgenstein1953}, that simple solution is no longer available, and one must wonder which features of \emph{dog} are included in the word meaning and which are part of some more general knowledge of the world. We cannot possibly resolve this question here. However, an advantage of our account (see next) is that word meanings are embedded in our knowledge of the world, such that hearing the word \emph{dog} gives one access to the knowledge necessary to understand the above interchange. Our discussion will not depend on any particular way of drawing the line between linguistic meaning and more general knowledge. NLP models often measure their success in tasks that likely go beyond the use of word meaning alone, such as answering multiple-choice questions about the meaning of a paragraph or text summarization. In evaluating such models, we will examine whether people and models understand words in the same way,  without worrying too much about the demarcation between word meaning and general knowledge.

In psychology, the predominant approach to word meaning is that it is a mapping of words onto conceptual structure \citep[see][ch. 11, for review]{Murphy2002}. That is, people have concepts that are the building blocks of their world knowledge, and the meaning of a word is essentially a pointer to some subpart of that knowledge. Concepts make the connection between language and the world argued for in philosophical and linguistic approaches to semantics, but in a psychologically plausible way. That is, when someone tells you, ``Watch out for the car,'' the word \emph{car} activates a concept, which contains information about what cars look like, enabling you to identify the object you're supposed to watch out for. The concept also contains information about why cars might be dangerous, what they do, and where they go. So, one plausible response to this warning would be to jump back onto the sidewalk, since you know that cars almost always drive on the road. Because a concept like dog is embedded in a network of other concepts and properties, when the word \emph{dog} occurs, it activates that concept which then gives access to all the other knowledge that one has about dogs, allowing Todd to interpret Marjorie's explanation. Thus, the concept that is pointed to by the word gives access to other information necessary for comprehension, even if it is not what many would consider part of word meaning.

Note that we are not proposing a learning theory in which conceptual structure exists first, and then words are just mapped onto it \citep[cf. discussion in][]{Lupyan2019}. Learning words can change one's concepts, and language is certainly a major way by which our knowledge of the world is increased. We discuss both below.

The conceptual approach to word meaning has two advantages as a psychological explanation over purely referential approaches. First, we know that people have concepts and knowledge of their world, so this is a plausible psychological representation, unlike infinite sets of objects in possible worlds. Second, in this account, words are connected to the world, because words are connected to concepts, which are in turn connected to the world, through perceptual and motor mechanisms. We use concepts to classify and think about objects in the world even when we are not talking about them. When we see a dog, that activates various perceptual representations that eventually activate our concept of dogs, which could then result in a verbal remark, like ``There's the dog,'' or ``I didn't know you have a dog,'' or an action, such as petting the animal. That is, use of the word \emph{dog} is causally connected to the presence of an actual dog. (The exact nature of this causal connection is a matter of debate among philosophers.) Of course, speakers and writers can discuss objects and situations that are not currently present, but the words used gain their meaning in part through their connections to concepts that are in fact linked to the world. When a speaker gives you new information through language, that changes your representation of the world and will potentially be useful to you later on.

There is considerable, though not universal \citep[][]{Lupyan2019}, agreement regarding the conceptual approach among researchers of word learning in particular, starting with seminal proposals of word learning by Eve Clark (\citeyear{Clark1983}) and Susan Carey (\citeyear{Carey1978}). Indeed, publications often vary between talking about word learning and concept learning as if they are the same thing. And, of course, they often are. Perhaps your child can correctly label cows when you go to visit a farm. However, you might also correct the child in some cases by saying, ``No, that's not a cow, it's a goat.'' By introducing a new word, \emph{goat}, you are encouraging your child to note the differences between that referent and the cows: smaller size, different proportions, longer head, possibly a beard, and so on, thereby forming a new concept. If you had never mentioned the word \emph{goat}, your child might not have formed the concept of goats and continued to include them in a broader category of cows. The requirement to use words the way adults do can act as a stimulus for children to distinguish objects in the world a certain way \citep{Mervis1987}. That is, word learning can influence one's concepts, because word learning involves learning what kinds of things there are in the world (see \citeauthor{Murphy2002}, \citeyear{Murphy2002}, for more discussion). 

Our focus is on the representation of words, but this does not mean that we will only consider words in isolation. The best way to identify whether a representation works is to see whether it can function properly in context---in reference to the world, or as part of a sentence or utterance. By focusing on words, we mean to exclude sentence- or discourse-level semantic phenomena, such as sentence ambiguity, identifying contradictions or oxymorons, sentence entailment, interactions among quantifiers, and similar phenomena that occupy semanticists. Those issues should be studied with respect to NLP models but are beyond the scope of the present investigation. Nonetheless, there will be no shortage of sentences in our examples as ways of illustrating how a model understands words.

\section{Desiderata for a model of psychological semantics} \label{sec_intro_desiderata}
We have discussed the theoretical basis for this view of meaning in some detail, because it is exactly this connection to world knowledge, the world, and language use that we argue is largely missing in current NLP systems. In order to be an adequate theory of psychological semantics, a proposed representation must provide the basis for carrying out a number of flexible behaviors---physical, verbal, and mental---that rely on conceptual representation as summarized in the list of five desiderata in Table \ref{table_desiderata}. These desiderata are meant to apply to \emph{any} theory of psychological semantics, computational or otherwise. We provide short examples of the entries here; the remainder of our article goes into further detail. We should emphasize that these desiderata are by no means exhaustive. Most of them relate to the most basic functions of language use that have been studied extensively in cognitive science. A model that can accomplish all those things might not talk and understand as people do, but a model that cannot do one or more of these things would fall far short.

Imagine you are at the dinner table with family. You might look at the table and say, ``That knife is dirty.'' To produce this description (Table \ref{table_desiderata}; \#1), you clearly must recognize some key objects---place settings, silverware, residual food crust, along with their relations---and retrieve the correct words to refer to them. You then drew attention to a property of one of them, in particular, that one knife is unsuitable for use. Someone listening might replace the offending knife with a new knife, thereby carrying out your (implicit) instruction (\#3). That person had to figure out what kind of thing you had in mind by saying ``knife'' and get that kind of thing instead of a spoon, plate, or other handy object. If you don't care for the kind of knife someone hands you, you might say, ``I was hoping for a butter knife.'' When you said that sentence, you had formed a goal of what kind of thing you wanted and then translated that idea into the English phrase ``butter knife'' (\#2) so that people would know what to get.

So, even with this simple interchange, we can see that words can be activated by perceptual input and by mental representations and that they in turn can cause others to form a particular representation and to take action in the world. Your listener must understand the conceptual combination ``butter knife'' as indicating a shorter, blunt knife specifically made for butter or cheese (\#4). Such phrases are constructed on the fly in English, and speakers make up and understand novel ones in everyday use. Finally, if you choose to, you might tell the children at the table, ``Forks go on the left, knives on the right,'' which is not a description of that particular table but rather information about knives and forks that you hope (vainly) the children will add to their store of knowledge about the world (\#5). If you are successful, this could some day result in the children placing forks on the left (i.e., changing the future world) when they set the table.

\begin{table}[t]
\captionsetup{
singlelinecheck=false}
\caption{\\Five desiderata. Word representations should support these basic functions of language use.}
\label{table_desiderata}
\begin{tabular}{ll}
\textbf{\underline{Behavior to be explained}} & \textbf{\underline{Examples}}\tabularnewline\tabularnewline

\begin{minipage}[t]{0.47\columnwidth}\raggedright
1. Describing a perceptually present scenario, or understanding such a
description.\strut
\end{minipage} & \begin{minipage}[t]{0.47\columnwidth}\raggedright
\emph{That knife is in the wrong place.}

\emph{The orangutan is using a makeshift umbrella.}\strut
\end{minipage}\tabularnewline\tabularnewline

\begin{minipage}[t]{0.47\columnwidth}\raggedright
2. Choosing words on the basis of internal desires, goals, or plans.\strut
\end{minipage} & \begin{minipage}[t]{0.47\columnwidth}\raggedright
\emph{I am looking for a knife to cut the butter.}

\emph{Book a flight from New York to Miami.}\strut
\end{minipage}\tabularnewline\tabularnewline

\begin{minipage}[t]{0.47\columnwidth}\raggedright
3. Responding to instructions and requests appropriately.\strut
\end{minipage} & \begin{minipage}[t]{0.47\columnwidth}\raggedright
\emph{Pick up the knife carefully.}

\emph{Find an object that is not the small ball.}\strut
\end{minipage}\tabularnewline\tabularnewline

\begin{minipage}[t]{0.47\columnwidth}\raggedright
4. Producing and understanding novel conceptual combinations.\strut
\end{minipage} & \begin{minipage}[t]{0.47\columnwidth}\raggedright
\emph{That's a real \underline{apartment dog}.}

\emph{The \underline{apple train} left the orchard.}\strut
\end{minipage}\tabularnewline\tabularnewline

\begin{minipage}[t]{0.47\columnwidth}\raggedright
5. Changing one's beliefs about the world based on linguistic input.\strut
\end{minipage} & \begin{minipage}[t]{0.47\columnwidth}\raggedright
\emph{Sharks are fish but dolphins are mammals.}

\emph{Umbrellas fail in winds over 18 knots.}\strut
\end{minipage}\tabularnewline



\end{tabular}
\end{table}

The first three desiderata are very basic functions of language. The final two are illustrative of the connection between language and knowledge. That is, language requires knowledge to be understood and also acts as a critical medium for increasing our knowledge. These two cases are just two of many possible examples.

\section{Computational Approaches to Word Meaning} \label{sec_NLP_models}
This now leads us to consider some typical examples of computational approaches to meaning, which will contrast greatly with what we have outlined above. The words \emph{semantics} and \emph{meaning} do not belong to anyone; there is no law saying that researchers in one field must use the words in the way another field dictates. Thus, when we point out these differences, we are not criticizing one or the other field for not conforming to psychological or linguistic usage. However, it is important to see what those differences are, so that there will not be confusion about which problems in ``semantics'' have been solved when the term is used differently by different researchers. If one proposes a theory of \emph{psychological} semantics, it must be the kind of theory that can do at least the things we described in Table \ref{table_desiderata}. Here we focus on approaches to word meaning that derive meanings from text relations, which have been described as theories of semantics by some of their proponents. 

\subsection{Word Representations}

A classic approach to word meaning is \emph{distributional semantics,} the idea that words have similar
meanings if they have similar patterns of usage and co-occurrence with other words \citep{Harris1954,Firth1957}. An influential text-based model based on these principles was the early work of Thomas Landauer and his colleagues in forming \emph{Latent Semantic Analysis} (LSA) \citep{Landauer1997} and the related system HAL \citep[\emph{HyperAnalog to Language};][]{Lund1996}. We will focus here on LSA, which was developed much more extensively. Although such models are no longer state-of-the-art, LSA is still commonly used in psycholinguistic research as a measure of relatedness of words or texts.\footnote{A Google Scholar search shows that the original \citet{Landauer1997} article was cited about 643 times in 2020. The citations seem to be both discussions of computational theories of meaning and papers that use LSA for evaluating experimental materials. Thus, although LSA is quite old by NLP standards, it is still influential and used in practice.} For example, in a priming experiment, you might try to ensure that the relatedness of primes and targets in different conditions is the same by showing that their similarities in LSA are about equal. LSA is also of interest because some of its proponents specifically argued that it is a model of human knowledge or word meaning \citep{Landauer2007}, and models using similar techniques continue to be tested against human psychological data \citep[e.g.,][]{Mandera2017a}. Such tests seem to imply that LSA and similar techniques might be a possible model of semantics. 

The original LSA model was trained on a corpus of 4.6 million words \citep[][p. 218]{Landauer1997}. To fit LSA, the corpus is divided into sections, called ``documents.'' Those sections might be articles (e.g., encyclopedia entries, newspaper stories) or simply paragraphs of a longer work. The system tags whether each word occurs in each document, forming a large matrix of words by documents. This matrix is reduced by singular value decomposition (SVD) to produce a low-dimensional vector for each word called a ``word embedding'' \citep[see][for a detailed explanation]{Martin2007}. This process has the effect of giving related words similar vectors, because their patterns of co-occurrence were similar in the original matrix. In LSA, it is not merely each word's co-occurrence with other words that is important, but second-order co-occurrences, such that if two words both co-occur with the same other words, their LSA meanings (vectors) will be similar. 

Thus, according to LSA, word meaning is represented through its word embedding. The vector is not interpretable in and of itself but in terms of its relation to other words. Word similarity is measured by calculating the cosine or dot product of the two vectors. Crude sentence representations can be constructed by adding together the vectors of its component words, or other operations analogous to predication \citep{Kintsch2001}. The system can be applied to various tests, such as choosing synonyms, completing a sentence, or evaluating whether a sentence is a good summary of a passage. Related probabilistic models can also identify topics in documents \citep{Blei2003,Griffiths2007a}. Finally, \citet{Landauer1997} showed that LSA could even score a ``passing'' grade on part of the TOEFL test of English for non-native speakers. 

Models based on text co-occurrence were limited in their ability to identify semantic relations beyond similarity (see below). However, a research program arose to augment such models so that they could serve as the basis for identifying superordinates, synonyms, part-whole relations, and the like \citep{Baroni2010b,Baroni2014a}. One technique was to start with labeled corpora, in which the part of speech of each word was identified, allowing better identification of relational vs. substantive terms. Another technique was to look for specific kinds of patterns that linguistic analysis suggests would indicate a given relation, for example, noun-verb-noun phrases, adjective-noun pairings, possessives, prepositional phrases, and so on. Such approaches were fairly successful in identifying specific lexical relations, but they required linguistic sophistication and specific analyses to identify each relation of interest, and the emphasis in the field seems to have returned to less directed models that rely on much more intensive computation, as we describe next.

After the development of LSA and related models, a different approach arose for deriving meaning from text sources \citep{Mikolov2013a,Pennington2014}. To distinguish these classes of methods, \citet{Baroni2014} called the earlier approach \emph{count models} (as they rely on co-occurrence counts of words) and the newer approach \emph{predictive models}. As the name suggests, these models attempt to learn words by trying to predict the probability $P(w|C)$ of a missing word $w$ given its context $C$ (alternatively, \emph{skip-gram models}, which we don't discuss here, predict a surrounding context given a word). For example, say the prompt is, ``Chris bit into the juicy \mask and placed it on the kitchen counter.'' A plausible guess would be some kind of food, possibly a fruit like plum or orange.

\begin{figure}[t]
	\centering
	\includegraphics[width=\linewidth]{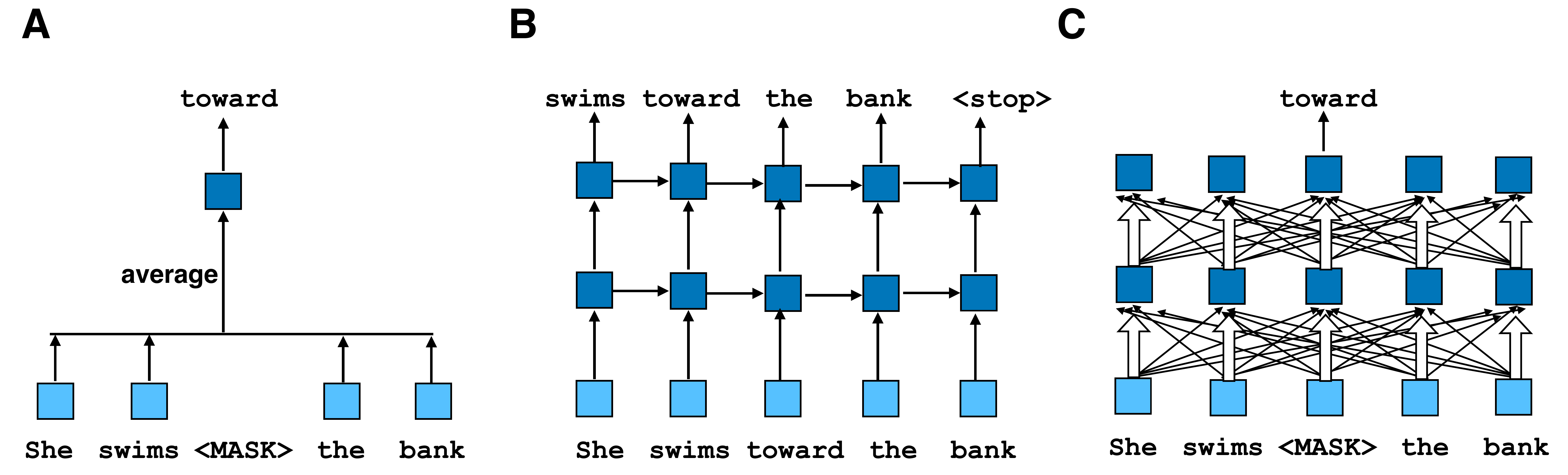}
	\caption{Popular neural net architectures that we analyze in this article, including CBOW (A), RNN (B), and BERT (C). Models (A) and (C) predict a missing word (``toward'') given its context (``She swims $<$MASK$>$ the bank''), while (B) predicts each word in the sentence given the previous words. Light blue boxes indicate word embeddings (vectors), and dark blue boxes indicate hidden embeddings (also vectors) after incorporating context. The hollow arrows in (C) are residual connections.}
    \label{fig_nlp_models}
\end{figure}

A popular predictive model is \emph{Continuous Bag-of-Words} \citep[CBOW; ][]{Mikolov2013a}, which is illustrated in Figure \ref{fig_nlp_models}A. CBOW has been trained on enormous corpora; for instance, in this article, we analyze a large-scale CBOW model trained on the Common Crawl corpus of 630 billion words. CBOW learns a word embedding for each word in the corpus (Figure \ref{fig_nlp_models}A; light blue boxes), which are the analogs of the LSA word embeddings. CBOW takes a context window $C$ and computes the average embedding, and then compares this average vector to possible output words $w$ using the equations, 
\begin{equation*} \label{eq_cbow}
\text{sim}(w,C) = w^{\intercal} (\frac{1}{|C|} \sum_{c_i \in C} c_i), \ \ \ \ \ P(w|C) = \frac{e^{\text{sim}(w,C)}}{\sum_{w'} e^{\text{sim}(w',C)}}. 
\end{equation*}
The $d$-dimensional embedding for each candidate output word is $w\in\mathbb{R}^d$, and the embedding for each word in the context window is $c_i\in\mathbb{R}^d$. In essence, the contextual window is summarized by the average of the word embeddings (Figure \ref{fig_nlp_models}A; dark blue box). Then, the similarities must be computed between each of the candidate words and the contextual summary (via dot product), using a softmax function to normalize these similarities to become the probabilities $P(w|C)$ used for prediction. The word embeddings are the main trainable CBOW model parameters (light blue boxes), and they are learned via gradient ascent to maximize the (approximate) log-probability of the masked word.

Averaging (or summing) word embeddings has also been studied as a means of composition. At a phrasal level, \citet{Mikolov2013a} added together word embeddings to construct phrase representations. For example, ``French'' + ``actress'' resulted in a vector most similar to ``Juliet Binoche'' (a prominent French actress); ``Vietnam'' + ``capital'' resulted in ``Hanoi.'' In other work, \citet{Baroni2010} studied adjective-noun compounds using LSA embeddings (``bad luck'' or ``important route''), finding that matrix multiplication was better than additive models at reconstructing the representation of adjective-noun phrases (with the adjective as a matrix and the noun as a vector). It is remarkable that useful phrase representations can be constructed in such simple ways, but building sentence representation is more complicated as we will see next. 

\subsection{Sentence representations}

Sentences specify particular relations among the entities and actions they describe, and
comprehension requires recovery of those relations. \emph{The angry dog bit a sleeping snake} does not mean the same as \emph{A sleeping dog bit the angry snake}. Jumbling the words usually destroys the semantic representation, though speakers may be able to figure out what meaning might have possibly been intended: \emph{sleeping angry bit snake dog the a}. However, if one derives sentence meaning by adding together the vectors of the words in a sentence, one will arrive at the identical representation for all of the above examples, sensible and nonsensical. A model aiming to build sentence representations from word representations would need to have a model of syntax and sentential semantics in order to combine the words to form propositions­---an extraordinarily difficult problem. Computing sentence representations by summing word embeddings, such as the word embeddings learned by LSA or CBOW, is a non-starter for capturing the full richness of sentence meaning.

More sophisticated predictive models---known as \emph{language models}---build sentence representations using neural networks \citep[e.g.,][]{Elman1990,Devlin2019,Radford2018a}. As with CBOW, language models learn representations that are useful for predicting missing words given their surrounding context. Although basic CBOW discards word order (see \citeauthor{Mikolov2018}, \citeyear{Mikolov2018}, for an extension that uses it), language models use word order to learn meaningful syntactic and semantic structure, to some degree.

Language models are more computationally intensive to train than mere word representations. To jumpstart learning, language models can be initialized with the pre-trained word embeddings from a simpler model (CBOW), which make up the first layer of the language model (Figure \ref{fig_nlp_models}B and C; light blue boxes). During training, these word embeddings are fine-tuned along with all of the other downstream parameters. With enough data and a sufficient network capacity, the hope is that a model trained to predict missing words will learn syntactic and semantic knowledge about language---at least enough to solve practical NLP problems.

In pioneering work, Jeffrey Elman (\citeyear{Elman1990}) showed that Recurrent Neural Networks (RNNs) can learn meaningful linguistic structure when trained to predict the next word in a sequence (predicting the next step/word in a time series given the past steps is known as \emph{autoregressive modeling}). As shown in Figure \ref{fig_nlp_models}B, RNNs achieve recurrence by using the previous hidden vectors as additional input when predicting the next word. Through this mechanism, the hidden representation of each word is influenced by the representation of previous words. (Figure \ref{fig_nlp_models}B shows an RNN with two layers. The hidden representations are the dark blue boxes above each word.) Elman showed that RNNs trained on simple artificial sentences can show emergent lexical classes---implicit in how the word embeddings cluster---such as nouns, transitive verbs, and intransitive verbs. Subsequent work introduced RNNs with more sophisticated gating and memory mechanisms, such as Long Short-Term Memory \citep{Hochreiter1997} or Gated Recurrent Units \citep{Cho2014}, allowing networks to store and retrieve information over longer time scales. For RNN-based language models, sentences can be summarized and passed to downstream processing through a variety of methods: extracting the last time step's hidden vector, computing a simple average over hidden vectors across all time steps, or computing a weighted average over hidden vectors using weights determined on-the-fly by a downstream process \citep[known as \emph{attention};][]{Bahdanau2015}.

A new architecture, the Transformer, has started to dominate the leaderboards for language modeling and other NLP tasks \citep{Vaswani2017}. A Transformer architecture is shown in Figure \ref{fig_nlp_models}C. Transformers are neural networks that operate on sets: a transformer layer takes a set of isolated embeddings as input (Figure \ref{fig_nlp_models}C; light blue) and produces a set of contextually-informed hidden embeddings (dark blue). Residual connections (large arrows in the figure) help preserve the identity of these word representations as they flow through each layer, meaning that each input element (word embedding) corresponds more strongly to one element in the output set (the transformed word embedding). As with RNNs, transformer layers can be stacked for deeper contextualization (Figure \ref{fig_nlp_models}C shows two layers). Unlike RNNs, Transformers use a ``self-attention'' mechanism that facilitates direct interaction between each element in the set with all other elements, without relying on an indirect recurrent pathway. 

The isolated input embeddings aim, as with CBOW, to represent the meaning of words in isolation (Figure \ref{fig_nlp_models}C; light blue).\footnote{While our discussion uses the term ``word embeddings'' for simplicity, more complex tokenizations based on pieces of words are typically used in large-scale systems \citep[e.g.,][]{Sennrich2016}.} As a first step, these input word embeddings are concatenated with positional information to mark word order. Through each transformer layer, the word embeddings are updated based on the other words in the sentence; for instance, a homonym like ``bank'' must, in some sense, have a word embedding that captures multiple meanings of the word in isolation (Figure \ref{fig_nlp_models}C; light blue box above ``bank''). When presented in context, such as ``She swims toward the \emph{bank},'' the hidden representation should resolve to mean \emph{river bank} rather than \emph{financial institution} (Figure \ref{fig_nlp_models}C; dark blue boxes above ``bank''). We examine this type of contextual resolution in more detail later in the article.

As with RNN language models, Transformers can be trained as autoregressive models that predict the next word in a sequence, leading to networks that can seamlessly generate text but incorporate context in only a unidirectional manner (left-to-right in English). Below we examine GPT-2, a massive autoregressive Transformer with 1.5 billion parameters \citep{Radford2018a}, trained on a corpus of 45 million web pages linked from Reddit. (The new, much larger GPT-3 from \citeauthor{Brown2020}, \citeyear{Brown2020} was not available for evaluation at the time of writing. We consider the implications of GPT-3 in the General Discussion.) Alternatively, Transformer-based language models can be trained on the Cloze task of predicting randomly masked words given their bidirectional context (as shown in Figure \ref{fig_nlp_models}C). We examine a popular model of this flavor called BERT, with 340 million parameters, trained on a corpus of 3.3 billion words that combines Wikipedia and a corpus of books \citep{Devlin2019}.

\subsection{NLP as a theory of semantics} \label{nlp_as_theory}

These large-scale neural networks have been remarkably successful in NLP. They certainly do things that
could only have been dreamed of 25 years ago, and they provide help in many tasks such as translation, summarization, question answering, and natural language inference. They have limitations (see below), but they are still a work in progress in a dynamically changing field. They will continue to improve. However, what is their status as a theory of psychological semantics?

The driving force in NLP is the development of more powerful models that accomplish specific tasks rather than hypotheses about semantics. In most cases, NLP papers do not make claims about the relation of their models to psychology or linguistics \citep[there are exceptions, e.g.,][]{Baroni2010b}. Rather, most NLP papers are motivated by applications. Interestingly, the people who have been most likely to claim that these models could be theories of psychological semantics seem to be psychologists. For example, \citet{Landauer2007} explicitly argues that LSA provides a theory of semantics \citep[see][for a more nuanced approach]{Kintsch2007}. Others have tested the ability of such models to explain human data \citep[e.g.,][]{Baroni2014,Mandera2017a,Louwerse2007,Hill2015,Marelli2017,Ettinger2019,Rogers2020,Lewis2019}. That is, the models are tested in the same way one would test a theory of lexical representation. To the extent that we see agreement between humans and models, one can take  this as evidence for the validity of these approaches as psychological models.
Whether or not modelers intend their models as psychological accounts, we believe it is important to explicitly outline the challenges of interpreting all these models as psychological theories. (Earlier critiques of this approach within psychology are discussed in Section \ref{sec_summary_desiderata}.)

To work up to our argument, we first discuss older computational theories that have also been proposed as representations of psychological semantics and whose shortcomings are well known. 

\section{Early Computational Theories of Psychological Semantics} \label{sec_early_theories}

Perhaps the first computational theory of meaning in psychology was provided by Charles Osgood and his colleagues \citep{Osgood1957}. Working within a behaviorist framework, Osgood did not have a vocabulary to talk about mental representations as later researchers would. Therefore, he attempted to operationalize semantics in terms of behavioral measurements, in particular, rating words on adjectival dimensions like fast-slow and happy-sad. Osgood did this for many different scales and then submitted the results to a factor analysis, which was (often) able to reduce the data to three orthogonal scales, which he called \emph{evaluative} (good-bad), \emph{potency} (strong-weak), and \emph{activity} (tense-relaxed). Words with similar values on these scales behaved similarly in certain tests; experimental manipulations had sensible effects on the words' values on those dimensions.

Other approaches followed some years later, when new techniques of psychological scaling were invented. The creation of multi-dimensional scaling and clustering algorithms allowed researchers to represent the similarity of stimuli in comprehensible terms \citep{Shepard1974}. Within semantic memory research, \citet{Rips1973} famously fit scaling solutions for a set of mammals and birds (separately) and showed that these scaling solutions helped to predict categorization performance. First people rated the similarity of all the pairs of stimuli (e.g., bear-mouse). Then these data were combined and reduced into a low-dimensional spatial representation that simultaneously represented the similarities of all the items at once. Such scaling solutions can be seen as semantic representations and can predict human performance. For example, the distance between an item and its category name (e.g., bear-mammal, penguin-bird) predicted how long it took subjects to classify items in a sentence evaluation task (e.g., true or false: ``All bears are mammals''; ``All birds are penguins'').

Scaling solutions are useful for predicting behaviors that require comparing items to one another (e.g., classification, memory confusions), because these models represent the overall similarity of the scaled items. However, scaling solutions and Osgood's proposal suffer from the same problem, namely that they do not include the critical information that people must know in order to actually use those words in normal language activities. For example, in the scaling solution for mammals, Rips et al. noted that their solution seemed to indicate two main dimensions: size of the animal and its predacity (i.e., was it a predator or prey?). However, those dimensions do not specifically pick out particular mammals well enough to identify them. That is, which mammal is a mid-sized moderately predacious one? Which is large but very much prey? People know hundreds of animals. In order to know when to use names like \emph{sheep, goat,} or \emph{cow}, language users must know what they look like, what they eat, where they live, how they move around, and many other facts. If you see a drawing of a lone sheep, you can immediately label it ``sheep,'' without the drawing indicating the animal's size and without seeing it being preyed on. Osgood's three dimensions also simply don't tell us what the word means. You could know a word's evaluation, potency, and activity to three decimal places, but you still wouldn't know whether the word is a noun or a verb, concrete or abstract, or even what semantic domain it was in (as \citeauthor{Osgood1957}, \citeyear{Osgood1957}, p. 323, acknowledged).

Scaling solutions of this sort are good at representing the \emph{similarity} of various concepts or words \emph{to one another}, but they simply do not contain the body of knowledge people have about those things that controls the use of those words. Furthermore, the dimensions discovered in the scaling solutions are typically ones that help to distinguish the stimuli as a whole but often do not include information that is essential to understanding a specific item. Distinguishing sheep from goats might require knowledge of the specific bodily shapes, proportions, and parts of the two creatures, most of which is missing from the low-dimensional space. Scaling solutions do not address the primary desiderata of a theory of semantics (Table \ref{table_desiderata}): describing a perceptually present scenario, explaining what listeners understand when they hear that description, or choosing words based on desires and goals. If you are thinking that you hope to see goats at the farm, perhaps generating a mental image of what you hope to see, that would not match the information in the multidimensional scaling or in Osgood's dimensions sufficiently to pick out the word ``goat'' instead of the names of other farm animals. To be fair to early researchers in this area, there literally was not enough computing power in the world to run the kinds of sophisticated and massive NLP models that exist today, nor was there the know-how needed to construct the deep learning models that are now ubiquitous.

The low-dimensional scaling solutions are precursors to LSA embeddings and other NLP techniques developed since the 1990s, facilitated by the availability of large text corpora and more powerful computers. Scaling solutions based on human similarity judgments can be laborious to produce, especially for a large number of items. The similarity matrix for N items requires N*(N-1)/2 entries, each of which is an average of multiple human ratings. LSA skips the tedious step of collecting lexical judgments and instead assumes that a lot of information can instead be acquired from the co-occurrence patterns of words in text corpora. We do not question this assumption; a vector of 400 values (or larger in recent models) can certainly contain much information. Our central question, however, is whether modern NLP models provide an account of a language's semantics, as the question is understood in either linguistics or psychology. Do these modern approaches go far enough in closing the gap between early scaling methods and our desiderata for a theory of semantics? 

\section{Semantic similarity} \label{sec_similarity}

A basic method for analyzing word embeddings is to look at their nearest neighbors, as reported in a number of articles by model proponents (see below). It would be ideal to be able to identify the semantic features underlying each word, which would tell us when the model would apply those words. But since these models do not have readily interpretable features, researchers have looked at the words that are similar to them to try to understand what the models think the words mean. The way to do this is to identify words that have similar embeddings, which are the semantic representations that control word use and understanding in these models. These similar words should be from the same semantic domain and hopefully from the same category, with overlapping features. Words that are merely associated or that have some other kind of relation, like part-whole, object-attribute, or complementary functions are generally not semantically similar \citep[in the sense of][]{Tversky1977} and so are not used in the same way. For example, one of us has trouble distinguishing SUVs from mini-vans and is apt to apply one name to the other kind of vehicle. But he still has a general idea of what these words refer to and can usually be understood even when he makes a mistake with one of them. However, if he instead confused ``SUV'' with ``wheel,'' one would have to say that he was very confused about at least one of those words---even though wheels are part of SUVs and have an obvious semantic relationship. All SUVs have wheels, but they are not similar to wheels; they are similar to mini-vans. A number of models have problems with just this kind of confusion \citep[see][for a full discussion]{Hill2015}, although we will show that more recent NLP models seem to do much better.

Consider the nearest neighbors of \emph{dog} reported by \citet{Dennis2007} in the LSA Handbook: \emph{barked, dogs, wagging, collie, leash, barking, lassie, kennel}, and \emph{wag}. (Readers may easily test the model with their own words at \url{http://lsa.colorado.edu/}) Of the nearest neighbors, one is an inflected form of \emph{dog}, four are actions, two are associated things, and two are subordinates. We find this list problematic. The subordinates (\emph{collie} and \emph{lassie}) are clearly similar in meaning to \emph{dog}. However, the actions are not. Actions are from an entirely different semantic domain with different semantic properties, such as whether they are extended in time or punctate, which do not apply to objects. \emph{Barking} and \emph{wagging} are certainly actions that dogs do, but these actions should not have highly similar semantic representations to dogs. Similarly, the associated objects like leashes and kennels are obviously related to dogs, but a dog is not similar to a kennel. Dogs are animals that live and breathe and reproduce; kennels are human-made structures made of metal and wood. The properties of dogs are not properties of kennels, and vice versa. The words semantically similar to \emph{dog} should have been names for other mid-sized, domesticated mammals, like \emph{cat}, and other canines, like \emph{wolf} and \emph{coyote}. Superordinates of \emph{dog}, like \emph{pet} and \emph{mammal} are similar in meaning but are not in the list. See \citet{Lund1996} for similar issues with HAL's neighbors.

LSA, like most NLP models, keeps inflectional and morphologically modified versions of words separate; that is, \emph{dog} and \emph{dogs} are two separate words, a consequence of analyzing text rather than linguistic entities such as morphemes and stems. The output doesn't always correctly identify such words as being highly similar, however. \emph{Computed} has a similarity value of only .35 to \emph{compute} (cosine similarity; retrieved from the LSA website). The word \emph{saddle} is more similar to \emph{horse} than \emph{horses} is (.91 vs. .83). This could be a matter of insufficient data to fully identify the representations of less frequent forms, but it could be that in fact \emph{compute} and \emph{computed} occur in slightly different contexts, and the model is correctly identifying that. The problem is that the two words are synonymous except for tense. Surely they are more similar in meaning than \emph{compute} is to \emph{valuation} and \emph{inventory}, which are rated as more similar. Thus, if the LSA representation is ``correct'' in distinguishing these two word forms because they do not occur in the same contexts, then it is incorrect in claiming that its vectors represent semantic similarity.\footnote{One way to resolve this issue is to recognize that these embeddings reflect many types of relations simultaneously, and that classifiers (or other downstream processing) are needed to distinguish how two words---which are similar via cosine---actually relate to one another. This approach can be effective for detecting hypernymy \citep{Roller2014} and other semantic relations \citep[part-whole, contradiction, cause-effect, etc.][]{Lu2019,Baroni2014a}. Relatedly, the high-dimensional embeddings can be projected along selected axes to create more interpretable dimensions \citep[size, dangerous, etc.,][]{Grand2018}. Still, the need for these methods implies that LSA and similar models alone, and as typically used in psycholinguistics, are insufficient in representing semantic similarity.}

More sophisticated models may organize their semantic representations differently, and thus we examined the nearest neighbors of more recent NLP models. We tested a CBOW system trained on a much larger corpus of 630 billion words \citep[fastText implementation;][]{Mikolov2018}. The nine nearest neighbors of ``dog'' according to CBOW are as follows: \emph{dogs} (0.85 cosine similarity)\emph{, puppy} (0.79)\emph{, pup} (0.77)\emph{, canine} (0.74)\emph{, pet} (0.73)\emph{, doggie} (0.73)\emph{, beagle} (0.72)\emph{, dachshund} (0.72\emph{),} and \emph{cat} (0.71).\footnote{We used the CBOW implementation trained on Common Crawl provided by the fastText library \citep{Mikolov2018}. The nearest neighbors were also lightly filtered to exclude tokens with punctuation like ``dog---''. The results seem to depend on the size of the corpus and model details. \citet[][p. 75]{Mandera2017a} report nearest neighbors of the word \emph{elephant} for a CBOW model trained on movie subtitles, and the results are much like the LSA results---a mixture of words related in various ways---unlike the CBOW results we report here.} While LSA included actions and objects as close associates, CBOW does not; it only includes domesticated mammals and doesn't stray into other semantic domains. The nearest neighbors strictly include inflectional and morphological variants (\emph{dogs} and \emph{doggie}), subordinates (\emph{beagle} and \emph{dachshund}), a superordinate (\emph{canine}), and other close semantic associates (\emph{puppy, pup, pet,} and \emph{cat}). This CBOW appears to be much more successful than LSA in computing semantic similarity---a result that aligns with past work \citep{Baroni2014}---although corpus size and selection may be a critical factor. \citet{Hill2015} tested models on a particularly challenging set of word pairs, some of which were highly associated but not similar. They generally found that both count and prediction models were overly influenced by word association when judging similarity. (See also \citeauthor{Lupyan2019}, \citeyear{Lupyan2019}, for examples of both successes and shortcomings of such models.)

CBOW may even outperform more sophisticated language models on semantic similarity, since it focuses solely on learning word representations rather than sentence representations. Nevertheless, we examined the word embeddings of two large-scale language models, BERT \citep{Devlin2019} and GPT-2 \citep{Radford2018a}, as implemented in the Huggingface Transformers library.\footnote{We used the largest available models, bert-large-uncased and gpt2-xl, from \url{https://huggingface.co/transformers/}. The nearest neighbors were lightly filtered to exclude repeated instances due to spacing differences, and tokens with punctuation and numbers.} As with CBOW, we found semantically coherent neighbors. The word embeddings were extracted from the first layer of both models (the ``embedding layer''; Figure \ref{fig_nlp_models}C; light blue), before the self-attention layers that mix the word representations together. The nine nearest neighbors of ``dog'' according to BERT are \emph{dogs} (0.67 cosine similarity), \emph{cat} (0.44), \emph{horse} (0.42), \emph{animal} (0.38), \emph{canine} (0.37), \emph{pig} (0.37), \emph{puppy} (0.37), \emph{bulldog} (0.37), and \emph{hound} (0.35). Like with CBOW, all of these neighbors are from the same semantic domain. The details are not always exactly correct; surely \emph{canine} and \emph{puppy} are more semantically similar to \emph{dog} than \emph{horse} is; \emph{pig} should not be as similar as \emph{bulldog} is, and other canine animals are missing. However the list manages to include only inflectional and morphological variants, superordinates, subordinates, and other animals. Similarly, the eleven nearest neighbors for ``dog'' according to GPT-2 are \emph{dogs} (0.7),  \emph{Dog} (0.65),  \emph{canine} (0.54),  \emph{Dogs} (0.50),  \emph{puppy} (0.46),  \emph{cat} (0.38),  \emph{animal} (0.37),  \emph{pet} (0.37),  \emph{horse} (0.35),   \emph{pup} (0.35), and  \emph{puppies} (0.35).  The list is similar to BERT's except that GPT-2 is case sensitive. Again, the model may not be picking up some details, as \emph{horse} appears before other canines. 

The more powerful NLP approaches also seem to better capture inflectional and morphological variants. Unlike LSA, the stronger NLP models find that the most similar word to \emph{horse} is its plural form \emph{horses} (this pattern is also found for \emph{dog, dolphin, knife, bank,} etc.).  Similarly \emph{compute} and \emph{computed} are close neighbors in these models but not in LSA. Thus, these more recent models seem to have escaped some of the shortcomings of the earliest count models, although it still should be mentioned that \emph{saddle} is more similar to \emph{horse} than many other mammals are (e.g., \emph{Thoroughbreds, mule, donkeys}, and \emph{greyhound} for CBOW; \emph{colt, thoroughbred, zebra}, and \emph{bull} for BERT). Overall, the models are now in the right semantic ballpark but have not yet gotten the details right.

A more difficult test of semantic representation involves homonyms, like \emph{bank} that have multiple meanings, e.g., a river bank or a financial institution. Homonyms are a challenge for word embeddings: These words have multiple meanings but have only one embedding to represent them. (See \citeauthor{Kintsch2007}, \citeyear{Kintsch2007}, for a good discussion of approaches to ambiguity in statistical models with a single meaning for each word.) Transformers, however, are not restricted to isolated word embeddings. They can incorporate context from much larger chunks of text, with hope that the hidden embeddings for \emph{bank} can be refined appropriately (Figure \ref{fig_nlp_models}C; dark blue; see also \citeauthor{Ethayarajh2020}, \citeyear{Ethayarajh2020}). Previous work has argued that Transformers are especially well-suited to resolve the meaning of homonyms through context \citep[word sense disambiguation;][]{Wiedemann2019,McClelland2019}, as any model of psychological semantics must be capable of doing. Here, we test some simple cases on homonym resolution with a Transformer.

To evaluate homonym resolution, we first presented the word ``bank'' in an ambiguous sentence that is consistent with either meaning, ``She sees the \emph{bank}.'' The top-layer hidden embedding of ``bank'' was extracted (Fig \ref{fig_nlp_models}D; top dark blue box above ``bank'') and compared with the embedding of related top-layer embeddings of other words (\emph{treasury, ATM, shore}, and \emph{beach}) given the same framing, e.g., ``She sees the \emph{treasury}'' or ``She sees the \emph{shore}.'' We can make only targeted comparisons between particular embeddings rather than evaluate all of a word's neighbors, since evaluating sentences is computationally expensive. The results of these targeted tests are summarized in Table \ref{table_sim_bank}. As with the isolated embeddings, BERT sees \emph{bank} in the ambiguous context as more similar to the financial words \emph{treasury} and \emph{ATM} than it does to words related to bodies of water, \emph{shore} and \emph{beach}. However, if the framing for \emph{bank} instead suggests an aquatic meaning, ``She swims toward the \emph{bank}'' (Table \ref{table_sim_bank}; row 2), the contextualized embedding for \emph{bank} is now more similar to the embeddings for \emph{shore} and \emph{beach} than it is to \emph{treasury} and \emph{ATM}. Finally, if the framing for \emph{bank} more clearly suggests the financial meaning---``She deposits it in the \emph{bank}''---the words \emph{treasury} and \emph{ATM} are again the most similar to \emph{bank} (Table \ref{table_sim_bank}; row 3). Thus, BERT seems promising as an architecture for resolving meaning given context, at least as measured through embedding similarity. Replacing the pronoun ``She'' in Table \ref{table_desiderata} to either ``He'' or ``I'' gave similar results. For this test of \emph{bank} disambiguation, we found that GPT-2 behaved similarly to BERT.

\begin{table}[t]
\centering
\caption{\\Cosine similarities between the top-level embedding of ``bank'' in context (rows) and underlined words in a neutral frame, ``She sees the \mask'' (columns) in the BERT model.}
\label{table_sim_bank}
\begin{tabular}{lllll}
& \multicolumn{4}{c}{Fillers for ``She sees the \mask.''}\tabularnewline
& \underline{Treasury} & \underline{ATM} & \underline{shore} & \underline{beach}\tabularnewline
``She sees the \underline{bank}.'' & \textbf{0.62} & \textbf{0.73} & 0.61 &
0.60\tabularnewline
``She swims toward the \underline{bank}.'' & 0.48 & 0.51 & \textbf{0.74} &
\textbf{0.63}\tabularnewline
``She deposits it at the \underline{bank}.'' & \textbf{0.55} & \textbf{0.66} &
0.51 & 0.53\tabularnewline
\end{tabular}

\vspace{1ex}
{\small \flushleft Note: The two highest rated word completions are in boldface. \par}
\end{table}

\begin{table}[t]
\centering
\caption{\\Cosine similarities between the top-level embedding of ``lead'' in context (rows) and words in a neutral frame, ``She has the \mask'' (columns) in BERT.}
\label{table_sim_lead}
\begin{tabular}{lllll}
& \multicolumn{4}{c}{Fillers for ``She has the \mask.''}\tabularnewline
& \underline{advantage} & \underline{win} & \underline{iron} & \underline{metal}\tabularnewline
``She has the \underline{lead}.'' & \textbf{0.55} & \textbf{0.59} & 0.54 &
0.54\tabularnewline
``She lifted the \underline{lead}.'' & 0.53 & 0.54 & \textbf{0.58} &
\textbf{0.59}\tabularnewline
``She scored and took the \underline{lead}.'' & \textbf{0.58} & \textbf{0.67} &
0.47 & 0.46\tabularnewline
\end{tabular}

\vspace{1ex}
{\small \flushleft Note: The two highest rated word completions are in boldface. \par}
\end{table}

To further examine BERT's abilities, we examined another homonym, ``lead,'' meaning an advantage versus a type of metal.\footnote{Actually,   this word is highly polysemous, covering six large pages in the OED.   We focus on these two common senses of the word, testing it in   sentence contexts that make clear it is a noun.} When presented in an ambiguous context (``She has the \emph{lead}.''), the embedding for \emph{lead} is slightly closer to meanings of \emph{advantage} and \emph{win} (Table \ref{table_sim_lead}; row 1).  However, when presented in a context that evokes interacting with a physical object or substance (``She lifted the \emph{lead}.''), the embedding for \emph{lead} is now more similar to \emph{iron} and \emph{metal} than to \emph{advantage} and \emph{win} (Table \ref{table_sim_lead}; row 2). In a context that evokes sports (``She scored and took the \emph{lead}.''), the embedding for \emph{lead} is now more similar to \emph{advantage} and \emph{win} than it is to the other words (Table \ref{table_sim_lead}; row 3). Unlike BERT, we found that GPT-2 preferred the \emph{advantage} interpretation in all cases. Taking these results and related studies together \citep{Ethayarajh2020,Wiedemann2019}, it's clear that BERT has some ability to resolve the meaning of homonyms given context.

We must note that this test is not extremely difficult, as it  shows that the word's representation in context shifts towards the correct meaning rather than actually representing the correct meaning in detail. A complete model would have to arrive at a specific representation of the word meaning in detail, e.g., a heavy, flexible metal, poisonous to consume, able to block radiation, etc., while rejecting the properties from other noun and verb senses. Moreover, simpler models can also disambiguate some polysemous words through their surrounding words, either by computing the average embedding of the neighbors \citep{Erk2012} or by inferring the latent topic \citep{Griffiths2007a}. It's notable, though, how natural the Transformer architecture is for disambiguating meanings, without using any auxiliary machinery or training.

Their imperfections notwithstanding, large-scale predictive models appear to be much improved compared to LSA. The NLP models can capture important aspects of semantic similarity, including that nearest neighbors should come from the same semantic domain and that word meanings can be resolved with context. LSA is still widely used in psychological studies, but these examinations and the work of others \citep{Baroni2014} speak to replacing LSA with stronger word embeddings from techniques such as CBOW, which are readily available in pre-trained form \citep[see fastText software;][]{Mikolov2018}. Next, we discuss more quantitative tests than the nearest neighbor analyses popular in this literature, as well as more direct comparisons with human judgments.

\subsection{Human Behavioral Judgments and Ratings}

Researchers have sometimes compared the judgments of NLP models to those of humans who rated pairs of items on various dimensions. Positive results for such comparisons would seem to give credence to the notion that the models represent meaning as people do. \citet{Pennington2014} compared GLoVE similarity ratings to human similarity and relatedness ratings, for example, finding that correlations ranged from 0.48 to 0.84 on a variety of datasets. More recently, \citet{Mandera2017a} tested various models on their abilities to predict word similarity and relatedness ratings. The authors note that a major test of lexical access is semantic priming, and so they also employed the models to predict the size of priming effects of word pairs from a large database of priming studies. The results are difficult to summarize, but one important generalization (p. 75) is that the large text corpora were best for vocabulary-type tests (like TOEFL), but smaller corpora based on film scripts and subtitles were quite good for association and priming. In general, predictive models did better than count models, consistent with findings from \citet{Baroni2014}. Given that some of these comparisons were performed by psychologists and appeared in psychological journals, this seems to imply that such models provide accounts of psychological semantics. For example, \citet[][p. 57]{Mandera2017a} describe their project as investigating ``the relevance of these models for psycholinguistic theories,'' and at the end conclude that ``we can unequivocally assert that distributional semantics can successfully explain semantic priming data'' (p. 75), although they do not draw specific psychological conclusions.

One important point is that that not all of these tests are providing windows onto word meaning per se. That's not to say that there is no influence of semantics, but some tests are primarily (or equally) subject to other variables. For example, consider word priming and association. Both of these are strongly affected by frequency and co-occurrence of words (and their referents). Indeed, for a long time it was a controversial question as to whether semantic similarity caused lexical priming at all when the words were not actually associated. In a meta-analysis, \citet{Lucas2000} found semantic similarity does cause priming, but less than association does (about half the effect of association priming). The problem with association is that it is often reliant on contiguity, and although similar things may certainly co-occur in life or in language---e.g., cats and dogs---so do things that are not similar, like cat and meow, cat and tuna, cat and whiskers, or cat and bowl. Knowing that ``cat'' is related to ``bowl'' and so on does not tell you what a cat is, what it looks like, or what things in the world should be called \emph{cat}. Thus, capturing word associations in a model does not necessarily mean that one has captured word meaning. If it is claimed that the cosines of word embeddings can predict \emph{both} similarity and associations (priming, etc.), this suggests a conflation of theoretical constructs that needs to be clarified, as these are not psychologically the same things. This also has implications for the use of LSA and other models for the purposes of experimental control: It must be carefully considered exactly what the similarity of word (or phrase) representations in a model is measuring (e.g., association? relatedness? similarity?) when such models are used to equate materials.

\section{Desiderata} \label{sec_main_desiderata}

So far, we have examined newer NLP models in terms of close semantic similarity and found the results to be promising. However, semantic similarity is only one component that a model of psychological semantics would need to explain. At the beginning of this article, we asked what human behaviors a theory of lexical semantics should explain. We suggested five desiderata (Table \ref{table_desiderata}). Next, we review these desiderata in context of recent achievements in NLP and AI more generally. We conclude that despite the expanding capabilities of modern systems, they are not yet plausible accounts of psychological semantics.

The following sections examine each of the desiderata in turn. In general, we find that the purely text-based NLP systems that are most cited in psycholinguistics research can be quickly shown to be inadequate. However, there are many recent models that are less well known to psychologists and that show much more promise. We review the most relevant such systems for each desideratum and also analyze where its shortcomings may still lie. Many of these models include interfaces with vision and action, allowing them to verbally describe or act on the world based on linguistic input. Because models have been extended in different ways, we consider different ones in each section. It is conceivable that eventually, such models could be linked up to create a complete system (or that one model could be integrated with different input and output modules). In fact, many problem-specific systems use the same pre-trained word embeddings or language models as their starting point.

It is important to emphasize that we are not requiring a model of semantics to actually interact with the world in all the ways we discuss below. However, a potential theory of semantics needs representations of objects, properties, relations, categories, and so on that could describe the world if appropriate interfaces were provided. A model can be a potential theory of human semantics even if it doesn't have all the elements necessary to actually perform the task under discussion. However, attaching input and output modules to the model provides one way of making sure it forms semantic representations that allow its use of language to make contact with the world.

\subsection{Word representations should support describing a perceptually present scenario, or understanding such a description.} \label{sec_describe_scenario}

We understand that none of the NLP models discussed so far has the interface to actually interact with the world: no cameras, microphones, mechanical hands, etc. More importantly, their word representations are only meaningful in relation to other words; perceptual features and actions associated with the word's referent are not represented. For example, the word embedding of \emph{knife} will be related to words that describe the shape, parts, and functions of knives (say \emph{blade} and \emph{sharp}), but those words are not perceptual features and are represented in terms of their relations to still other words, and so on. As a result, there is nothing in the word vectors of text-based models that would allow their user to label an object or describe a scene. The visual features of a scene have no links to semantic features of the words. \citet{Bender2020} argue that text-only models cannot, even in principle, learn meaning \citep{Harnad1990}. In this regard, popular models of NLP are clearly inadequate as models of word meaning. We should note that this issue is not confined to computational models of semantics, as it has been a point of dispute about psychological theories of concepts and symbols, as represented by the debate over embodied cognition (see Section \ref{sec:past-critiques}). The essence of that debate is whether traditional theories of cognition can explain how mental symbols are connected to perceptual representations \citep{Barsalou1999}. Thus, this question has importance beyond the present discussion of NLP models.

The limitations of being based solely on text does not mean that NLP models will never serve as accounts of psychological semantics, however. There is a research area that combines computer vision with NLP models in ways that have greater promise for developing a more realistic model \citep[see][for a review]{Baroni2016}. \citet{Kiela2016} propose embodiment in virtual environments as a long-term research strategy for AI. Reviewing evidence from cognitive neuroscience, \citet{McClelland2019} argue that language representations are deeply integrated with multi-sensory perceptual representations, as well as representations of situations and events. They propose that language models should be placed into the context of other modules, of perception, action, memory, etc. \citet{Bisk2020} describe a roadmap for NLP that incorporates multi-modal, embodiment, and social factors. In this section, we follow previous work in highlighting the multi-modal nature of psychological semantics, while also discussing how its conceptual basis may need to go beyond current work in multi-modal machine learning.

Psychological semantics supports far more than word similarity. As we have noted, for human speakers the word meaning of \emph{knife} contains information about the shape, parts, functions, and likely locations of knives, such that when they see or think about something having these properties, they can produce the word \emph{knife}. A complete model would also understand the uses and implications of the concept (e.g., sharpness, dangerousness). If they hear, ``I got the knife, but the blade was dull,'' they should understand that the blade is part of the knife, and the blade is supposed to be sharp for carrying out certain functions (slicing, dicing, etc.). Similarly, if they hear, ``I got the knife, but the handle was broken,'' they should understand that the knife will be more difficult to use but is still potentially dangerous. Finally, if they hear ``I didn't have a knife, so I grabbed my keys to open the packing tape,'' they understand that the concept of a knife is characterized, in part, by a functional role that other objects can satisfy in some circumstances. A model must represent the parts and properties of knives in a coherent fashion in order to understand events described in text. Not everything
one knows about knives must be included in the lexical representation, but enough must be so that basic sentences can be understood and appropriate inferences drawn.

AI researchers are certainly working on various forms of multi-modal learning. A recent flurry of work has focused on integrating vision and language, leading to creative combinations of computer vision and NLP models. Active research areas include image caption generation \citep{Chen2015,Vinyals2014,Xu2015}, visual question answering \citep{Johnson2017a,Agrawal2017,Das2018}, visual question asking \citep{Mostafazadeh2016,Rothe2017,Wang2019}, zero-shot visual category learning \citep{Lazaridou2015,Xian2017}, and instruction following \citep{Hill2020,Ruis2020}. The multi-modal nature of these tasks grounds the word representations acquired by these models, as we discuss below.

\begin{figure}[t]
	\centering
	\includegraphics[width=.8\linewidth]{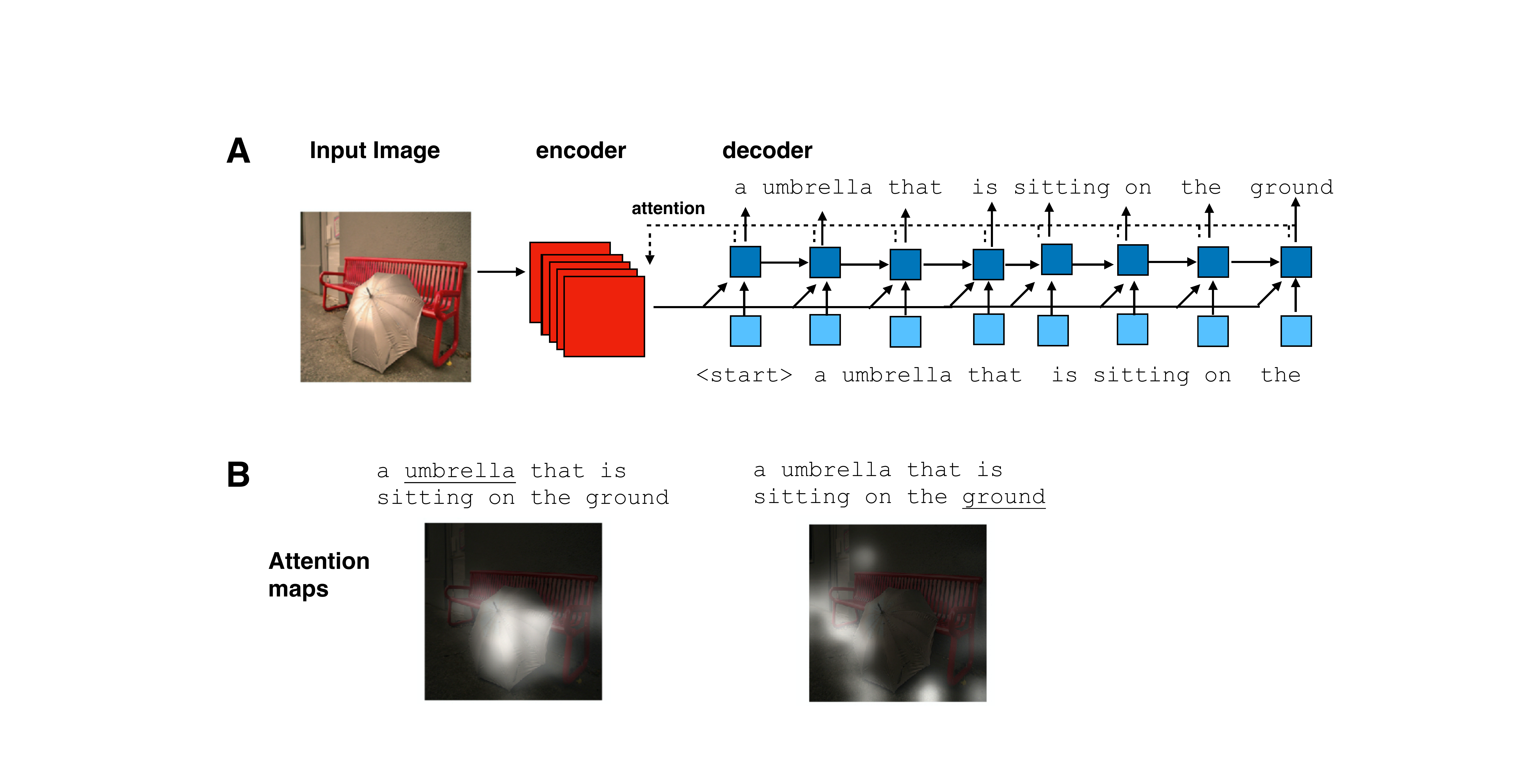}
	\caption{A neural architecture for caption generation \citep{Xu2015}.  (A) An input image is processed with a ConvNet encoder, producing a visual embedding for each spatial location (red). The encoder passes these messages to the recurrent decoder (blue), which produces a caption word-by-word. Each decoder step attends to different spatial locations in the input image. (B) Where the decoder is attending when producing words umbrella and ground as outputs.}
	\label{fig_arch1}
\end{figure}

Neural architectures for these tasks typically follow one of two templates. The first template is appropriate for tasks that take visual input and produce language output, such as caption generation or question asking. As shown in Figure \ref{fig_arch1}, the basic architecture involves two neural networks working together: a visual encoder and a language decoder. These models often start with a pre-trained encoder and, less frequently, a pre-trained decoder. The encoder is a convolutional neural network (ConvNet) pre-trained on an object recognition task, such as ImageNet.\footnote{ImageNet and all datasets in this article were used only for non-commercial research purposes, and not for training networks deployed in production or for other commercial uses.} The encoder produces a ``visual embedding'' (analogous to the word embeddings discussed earlier), and this is passed as a message to the decoder. The decoder is a language model that generates text, following the RNN language models discussed previously (Figure \ref{fig_nlp_models}B). The language decoder can be trained from scratch, or it can start with pre-trained word embeddings (e.g., CBOW) or a fully pre-trained language model (e.g., GPT-2). After initialization, the encoder and decoder are trained jointly (end-to-end) on the downstream task of interest, such as image captioning, allowing the visual embeddings to link up with the word embeddings in service of solving the task. The encoder can communicate with the decoder by passing a single visual embedding that summarizes the image content \citep{Vinyals2014}. More powerful models pass a set of visual embeddings from the encoder to the decoder, using different embeddings for different spatial locations in the image. Using these localized embeddings, the decoder learns to attend to different parts of the image as it produces words \citep{Xu2015}. Impressively, these models can show emergent visual-language alignment; the decoder often attends to the umbrella part of the image when producing the word ``umbrella'' in the caption (Figure \ref{fig_arch1}), perhaps analogous to human attention when describing scenes \citep{Griffin2000}.

\begin{figure}[t]
	\centering
	\includegraphics[width=.65\linewidth]{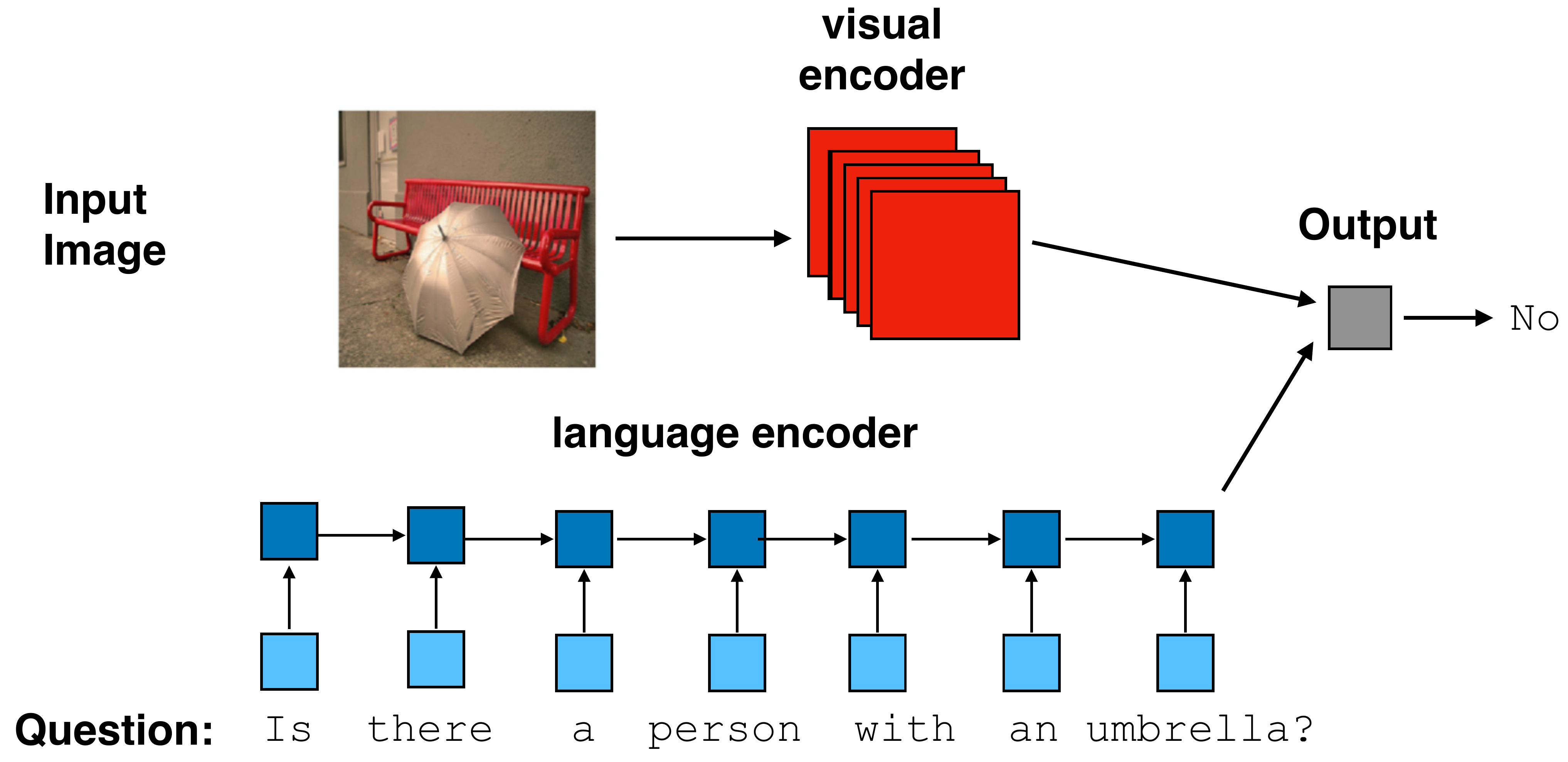}
	\caption{A neural architecture for visual question answering. An input image is processed with a ConvNet encoder (red), while a question input is processed with a RNN encoder (blue). Information from both encoders are combined in another network (gray) to produce the answer to the question.}
    \label{fig_arch2}
\end{figure}

The second template for neural architectures is common in grounded language understanding tasks such as question answering and instruction following. As shown in Figure \ref{fig_arch2}, the architecture has two encoders; a visual encoder processes the image (as in Figure \ref{fig_arch1}) and a language encoder processes the question or instruction. The encoders feed into another neural network that produces an output (e.g., the answer to the question, or the actions to perform a command). Similarly, the challenge is aligning the visual and language embeddings to successfully perform the task of interest.

These multi-modal models can learn useful visual-language alignments, as measured through correlational analyses of multi-modal embeddings \citep{Roads2020} and better-than-chance predictions of category labels for novel visual classes based on text-only experience with those labels \citep[zero-shot learning;][]{Xian2017}. Through pre-training and fine-tuning, these multi-modal neural networks can absorb truly immense quantities of visual and language experience. In a typical case, the models learn from a million or more labeled images for visual encoder pre-training, billions of words through word embedding pre-training, and hundreds of thousands of image-caption pairs for task fine-tuning over the entire model. Taken together, could multi-modal models trained on these massive datasets provide that necessary connection between words and the world?

\begin{figure}[!ht]
	\centering
	\includegraphics[width=.7\linewidth]{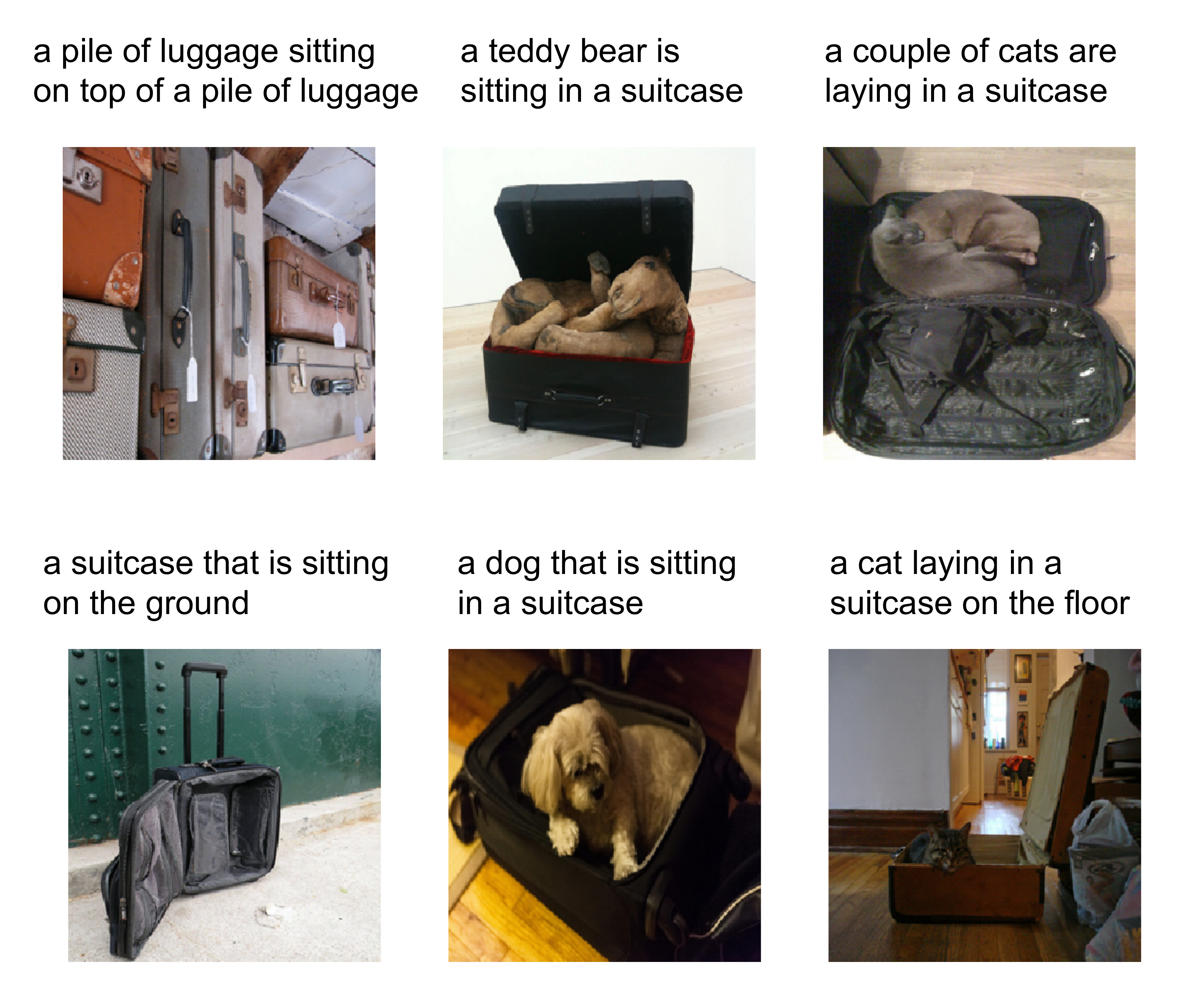}
	\caption{Captions generated by a neural network for scenes depicting suitcases.}
    \label{fig_suitcase}
\end{figure}

\begin{figure}[!ht]
	\centering
	\includegraphics[width=.95\linewidth]{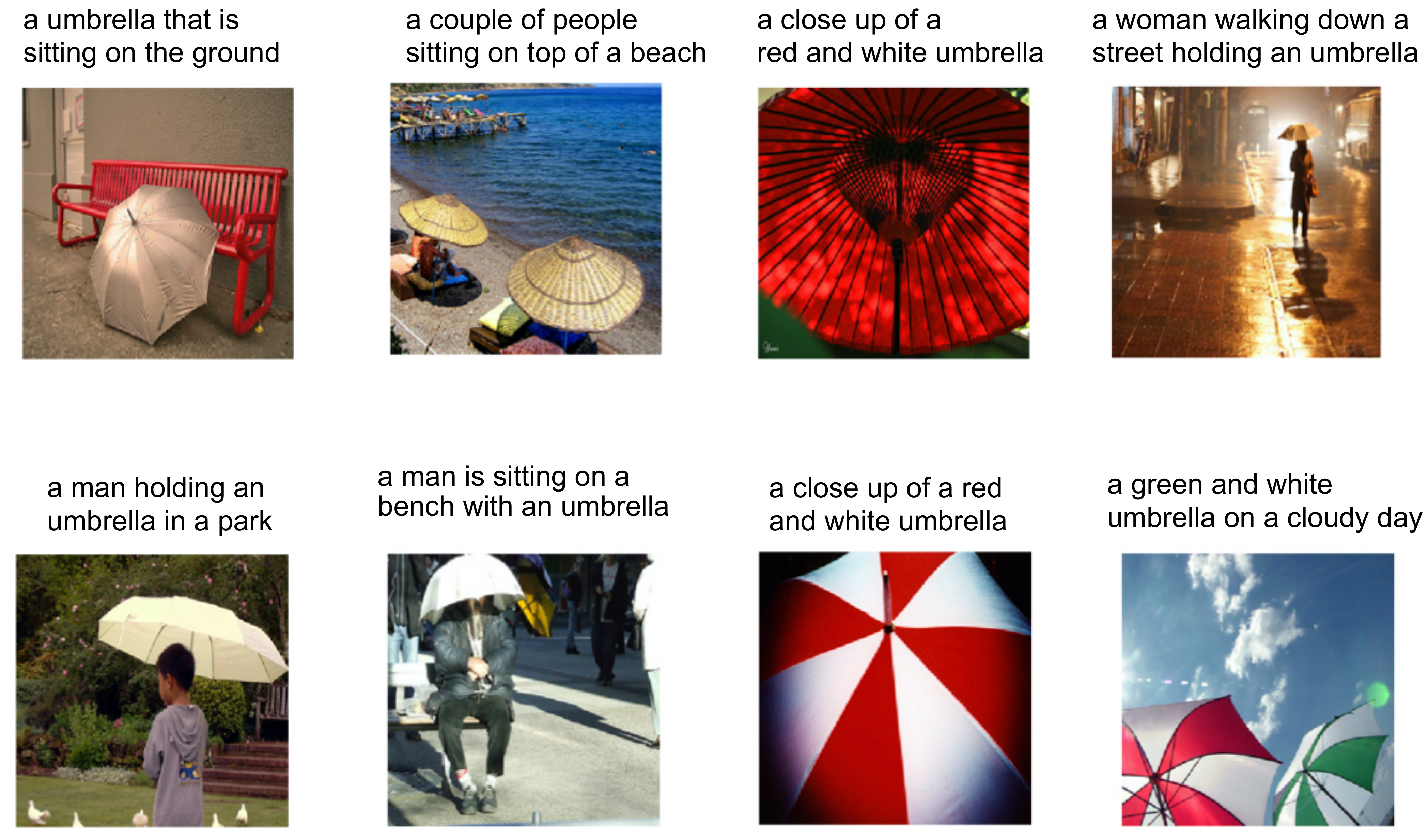}
	\caption{Captions generated by a neural network for scenes depicting umbrellas.}
    \label{fig_umbrellas}
\end{figure}

Multi-modal models help to provide substance to semantic representations, and this promising research direction will undoubtedly progress further. Nevertheless, as they stand these models are not satisfactory accounts of psychological semantics; learning to associate visual patterns with words is not sufficient to provide semantic knowledge. Returning to the knife, one must ultimately know things like the relation between the handle and the blade, what their names are, what each is used for, where and when a knife typically occurs, the function of a knife, how that function would change with a dull blade or a broken handle, and so on. This information must be abstract enough to generalize across modalities and be integrated with broader knowledge \citep{Murphy1985,Rumelhart1980}. For example, if Todd says, ``I need to open this box somehow,'' Marjorie could answer, ``Do you want a knife?'' General knowledge of packaging, cutting, and the functions of various tools can lead to retrieval of the concept and then the word \emph{knife}. By their nature, images do not directly represent functions and goals.

Of course, a proponent of NLP systems might propose that in fact such models do include detailed information about the structure of words, since that information is embodied in the huge text corpus that the model is trained on. There may well be sentences in a corpus about a knife being broken or the handle coming off or a dull knife being dangerous, the \emph{blade} in particular being dull, etc., which together might contain exactly the information needed to recognize and think about knives. If these models are paired with computer vision modules, they might well be able to create componential representations about the functions, properties, and parts of knives that are necessary for word use.

We focus on the concepts \emph{suitcase} and \emph{umbrella} as informal tests of this idea. We could have chosen a number of concepts to test instead (bird, car, bridge, etc.); we chose suitcase and umbrella in part because they are well-represented classes in the popular MSCOCO dataset of images labeled with captions \citep{Lin2014}, which focuses on  91 object categories. We tested a strong image captioning  system based on \citet{Xu2015} that was trained on MSCOCO and combines visual input with language processing in the way we have described (Figure \ref{fig_arch1}).\footnote{We used an updated version of \citet{Xu2015} for caption generation, available at \url{https://github.com/sgrvinod/a-PyTorch-Tutorial-to-Image-Captioning}} The training set included roughly 1600 scenes with suitcases and 2750 scenes with umbrellas, where each training scene was paired with five human-generated captions as supervision. Indeed, the trained network is quite good at generating captions for novel images like those it was trained on. As shown in Figures \ref{fig_suitcase} and \ref{fig_umbrellas}, the image captioning system succeeds at identifying salient suitcases and umbrellas, generating reasonable captions that mention these classes. Some of the captions demonstrate impressive acuity: ``A couple of cats are laying in a suitcase'' or ``A man is sitting on a bench with an umbrella.'' Additionally, NLP systems with word embeddings have a reasonable sense of what concepts are similar to suitcase and umbrella: The nearest neighbors of suitcase in CBOW are \emph{suitcases} (0.86), \emph{luggage} (0.73), \emph{duffel} (0.7), \emph{backpack} (0.67), and \emph{briefcase (0.67)}, and the neighbors of umbrella are \emph{umbrellas} (0.7), \emph{parasol} (0.59), \emph{parapluie} (0.57), \emph{brolly} (0.57), and \emph{raincoat} (0.54).\footnote{These are the five nearest neighbors that aren't misspellings or alternative capitalizations.} Considering these apparent successes, do these systems actually understand the meaning of \emph{suitcase} and \emph{umbrella}?

\begin{figure}[!t]
	\centering
	\includegraphics[width=.8\linewidth]{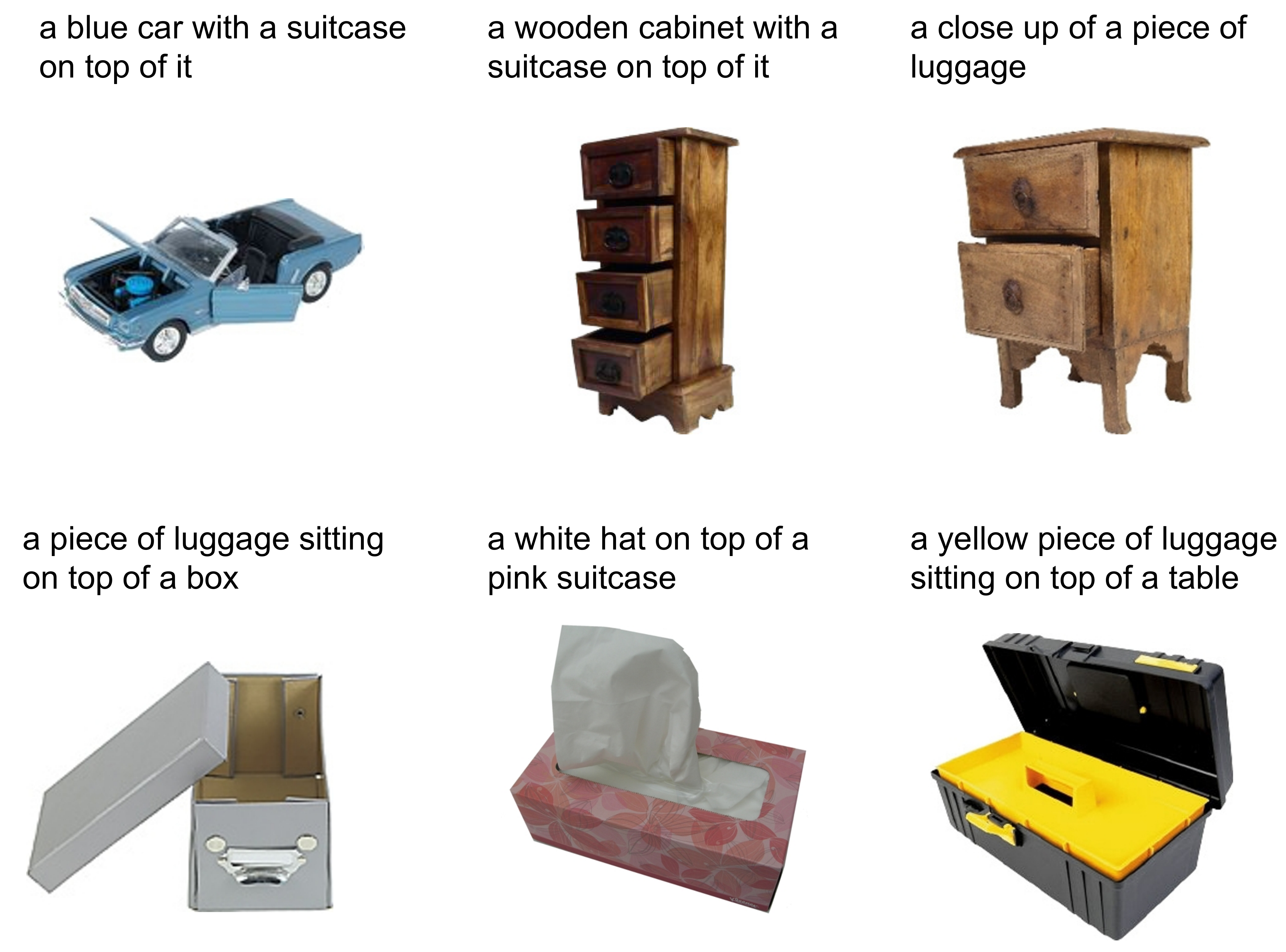}
	\caption{A caption generation system is fooled by images that share superficial features with the \emph{suitcases} and \emph{luggage} seen during training. Images from \citet{Brady2008,Brady2013a}.}
    \label{fig_suitcase_lure}
\end{figure}

\begin{figure}[!t]
	\centering
	\includegraphics[width=.8\linewidth]{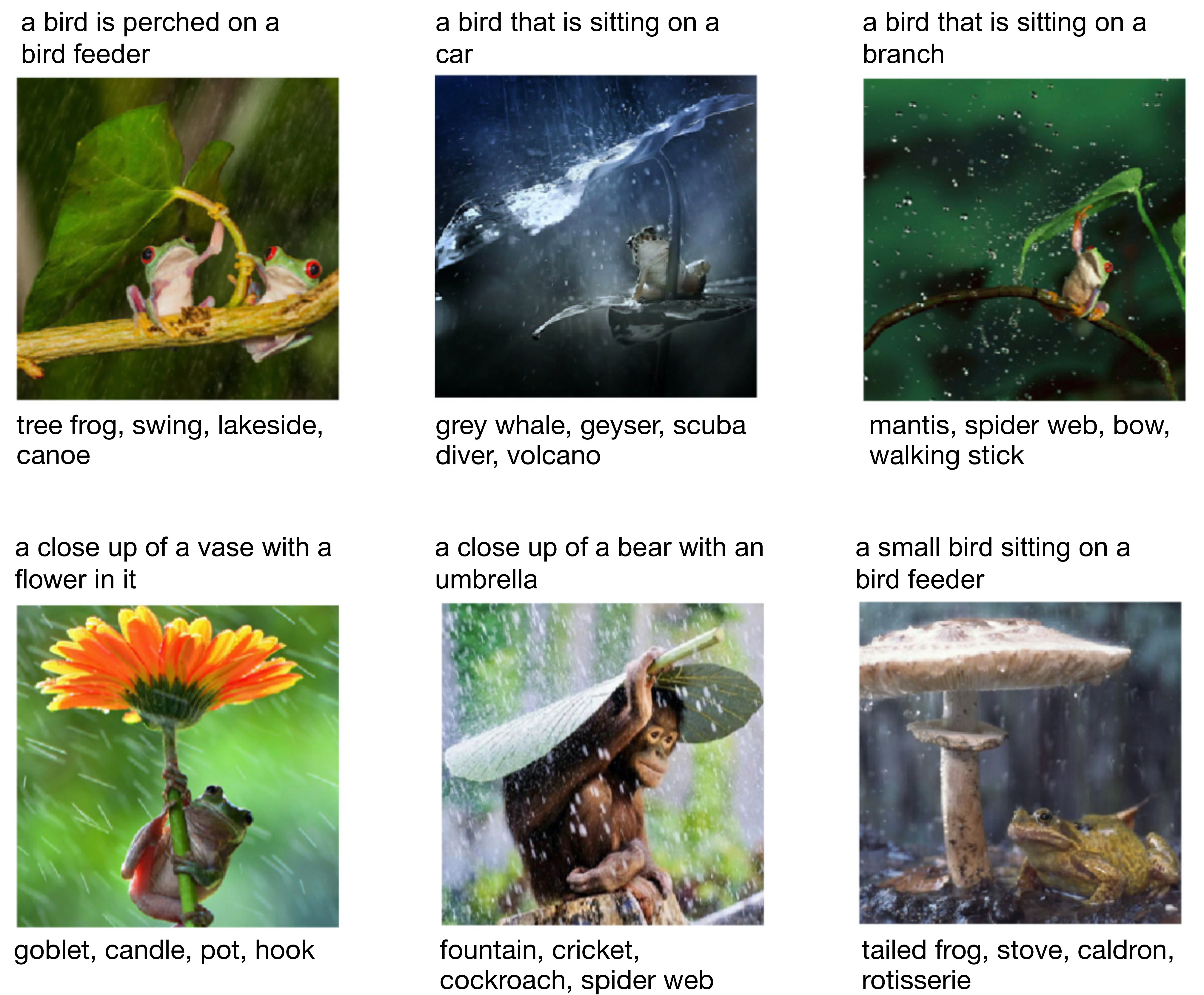}
	\caption{Captions generated by a neural network for scenes depicting natural umbrellas (caption shown above each image). Below each scene are four category labels that an object recognition system ranks higher than ``umbrella.'' Images reprinted with permission from Andrew Suryono and Edwin Giesbers.}
    \label{fig_natural_umbrellas}
\end{figure}

These vision-meets-language models can identify suitcases or umbrellas of the sort in their training set, but such tests do not require the detailed knowledge of objects that we discussed above, i.e., their parts, their functions, and why they have the structure they do. For instance, we find that the image captioning system overgeneralizes the words \emph{suitcase} and \emph{luggage} to images that share superficial features with these classes (Figure \ref{fig_suitcase_lure}). A car with an open hood shares some features with an open suitcase, but the neural network sees a blue car with a hallucinated ``suitcase on top of it,'' demonstrating a shaky understanding of both suitcases and cars. (\emph{Cars} are another featured class in the MSCOCO training set.) A dresser with open drawers superficially resembles the types of luggage piles present in the training distribution (Figure \ref{fig_suitcase}), but it is certainly not a ``a wooden cabinet with a suitcase on top of it'' or a ``close up of a piece of luggage.'' Indeed, confusing a wooden drawer or an entire dresser with a suitcase would be a very heavy and impractical mistake.

We probe understanding of umbrellas in a different way, using an ``ad hoc umbrella test.'' Since the image captioning system is good at identifying umbrellas in the types of images it was trained on, we evaluate whether it can generalize from this knowledge to more abstract meanings of \emph{umbrella} that are not directly reflected in the training corpus. In particular, some animals such as frogs and monkeys occasionally hold leaves, flowers, etc. as natural umbrellas when it is raining. Figure \ref{fig_natural_umbrellas} shows a range of natural umbrellas. These photos are striking because they are unexpected yet easily recognizable portrayals of an umbrella, or at least the function of an umbrella. In contrast, the state-of-the-art in machine understanding lacks these abstractions. A strong object recognition system \citep[ResNet50; ][]{He2016} that perfectly identifies the umbrellas in Figure \ref{fig_umbrellas} does not successfully classify any of the ad hoc umbrellas; in fact, the model ranks the label ``umbrella'' as quite unlikely compared to other labels. (The median rank for ``umbrella'' is 181 of 1000. The highest rank is 42.)\footnote{We tested the ResNet50 on umbrellas but not suitcases because only the former is a class in ImageNet ILSVRC challenge \citep{Deng2009}}.

Perhaps readers looking at those images may think,``If I saw these, I’m not sure I would spontaneously call them `umbrella.{'}'' Perhaps, but the task given to the ResNet50 is not quite spontaneous labeling---it is more like comparing the image to its entire vocabulary and seeing which words are most compatible with it. Therefore, we provide underneath each picture four labels that the ResNet ranks higher than umbrella. The first one is the highest ranked name. As can be seen, two of these names are very reasonable (e.g., ``tree frog'') but the others are rather far off (e.g., ``goblet, grey whale''). The three names following are all labels that are ranked higher than ``umbrella'' for that picture. As can be seen, many of them cannot apply literally or metaphorically to these scenes, such as ``fountain, rotisserie, hook, stove, walking stick,'' or ``volcano.'' Readers may judge for themselves whether they find the word \emph{umbrella} a more appropriate label than these. Our judgment is that it is much better, yet the ResNet50 model is not able to identify the relationships among the objects that make the scene umbrella-like more than hook-like or volcano-like.

The image captioning system \citep{Xu2015} does not fare much better than the object recognition system. Despite generating plausible captions related to umbrellas in Figure \ref{fig_umbrellas}, the caption system produces largely inaccurate descriptions of these scenes, including ``A bird that is sitting on a car'' and ``A close up of a vase with a flower in it'' (the phrases above the pictures in Figure \ref{fig_natural_umbrellas}). The novel combination of a natural object used as an umbrella for an unexpected animal seems to have disrupted the system's identification of the animals. It does use the word ``umbrella'' once, but notably the object recognizer did not consider ``umbrella'' a plausible label for this same photo, highlighting the issues with abstraction and robustness. A person may not spontaneously describe each image in Figure \ref{fig_natural_umbrellas} as an animal using a  umbrella, but surely they would recognize the fittingness of the description if asked directly.

Human word understanding is no doubt related to pattern recognition, but it is also conceptual and model-based, reflecting our understanding of the world around us \citep{Lake2016}. A computer vision or NLP algorithm learns patterns that distinguish \emph{umbrellas} from other entities such as \emph{suitcases} or \emph{cars}, while a person also learns \emph{simple models} of these concepts that cover their key parts, relations, uses, and ideals \citep{Rumelhart1980}. These models go on to support a variety of tasks: classifying and generating new examples, understanding parts and relations, inferring hidden properties, forming explanations, creating new yet related concepts, etc. \citep{LakeScience2015,Murphy2002,Solomon1999} and understanding sentences describing all those things. For example, when children learn about novel kinds of artifacts, they notice not only their shapes and parts, but also how those parts relate to the object's function. If the object is modified so that it is still similar to learned examples but misses a part that is essential to the function, children will not label it with the learned name \citep{KemlerNelson1995}. Object recognition is one part of the input to our semantic system, but there must also be connections to deeper knowledge of the type discussed by research on human concepts if we hope to capture people's labeling of the world \citep[e.g.,][]{Gelman2003,Keil1989,Murphy1988}.

In sum, recent combinations of computer vision and NLP models have taken important steps towards grounding text-based representations, an essential quality of any model of psychological semantics. These multi-modal models can accurately describe some perceptual scenarios and understand enough to answer certain questions. They are not yet at human levels of performance, likely because they lack conceptual knowledge to interpret the visual images they are trained on. Further progress may come from improvements in the training data; it's possible that through training on large-scale video/audio or richer 3D environmental simulations, models will come to develop more complete and useful semantic representations for words. However, our suggestion is that in order to explain how people describe and understand perceptual scenarios, it will be necessary to use more detailed and sophisticated representations of objects and scenes, of the sort provided by ``neuro-symbolic'' models that use structured representations (graphs, programs, etc.) to bridge between modalities rather than relying exclusively on neural mappings \citep{Johnson2017,Mao2019,Wang2019}. This is a promising direction for building richer models of psychological semantics, although current approaches aren't as developed as their neural mapping counterparts. However, from a broader perspective, we conclude that attempts to integrate knowledge of words with perceptual input are a promising avenue for developing word meanings that have true semantic content.

\subsection{Word representations should support choosing words on the basis of internal desires, goals, or plans.}

We do not walk around our world labeling the objects we see and hear. Instead, we tend to point out objects that are not there but should be, errors, unusual situations, goal states that we would like to take place in the future, and facts that we want our interlocutor to know. In some ways, this issue is the opposite of the one just discussed. People need to be able to produce words based on external inputs (describing a scene or labeling an object) but also based on internal ideas. Standard models of language production \citep{Levelt1993} propose that a sentence begins as a thought, a proposition of some kind that the speaker wishes to communicate. Words are matched to components of the thought, and a syntactic frame is selected that will match the structure of the proposition. Further processes spell out the details of the sentence, the phonetic structure of the words, and the motor sequences involved in producing them. A theory of word meaning must deal with the initial step in this process, the translation of idea to words.

The text-based systems we discuss do not have representations of ideas, per se. That is, the nonlinguistic thoughts that then generate linguistic representations are not present in them; all their representations are in terms of how words relate to other words. Since thought is not (or not solely) word-based \citep{Fodor1975,Murphy2002}, words must connect to concepts more generally. So far, this aspect is missing from the systems we have described, and so they cannot readily serve as models of spontaneous language production and conversation.

Of course, there are dialogue systems that can have conversations with users, and these systems vary in the structure and richness of their internal states. Current systems tend to fall into two types \citep{Chen2017}: 1) goal-directed systems with richer internal states but limited language skills, or 2) text-driven systems with more sophisticated language skills but limited internal states. Goal-directed dialogue systems often assist a user with specific tasks like making travel plans or restaurant reservations. Such systems do often have knowledge of their limited domains (e.g., flight schedules and costs, travel rules, typical preferences, etc.), and they choose their words on the basis of achieving goals and satisfying the user. In a theoretical sense, these systems may be considered to be semantic systems, in that their words connect to actual entities, actions, and events in the world. If a customer says, ``Let's book a morning flight from New York to Miami,'' the dialogue system may first translate the natural language utterance into a formal description specifying entities and relations through the process of semantic parsing \citep{Eisenstein2019}. Ultimately, a ticket may be sold involving a flight that actually will go from New York to Miami. The semantics of such systems are however limited by the scope of their applications. The flight-booking system knows about selling tickets in order to book a reservation, but they don't know about the paper instantiation of tickets or the visual properties of airplanes.

Goal-directed systems are a step forward in imbuing word representations with meaningful semantics; unfortunately, in other ways, these systems do not have sophisticated word representations. For language production, these systems typically serve up canned chunks of text \citep[even more recent neural-network-based dialogue systems do so too,][]{Bordes2016}, prioritizing the usefulness of the interaction rather than understanding the words it is saying. They lack the graded word representations that capture useful aspects of word similarity, as discussed earlier. For language understanding, many of these systems rely on hand-crafted features that map specific words (\emph{flight, connection, layover, price, tax, round-trip}, etc.) onto internal entities \citep{Young2013}, although other systems are beginning to integrate neural language understanding \citep{Bordes2016}. By strictly limiting the world in which such a model operates, it is able to manipulate real events and objects within that world. However, such models lack the linguistic flexibility and sophistication that allows people to name novel objects, make and understand novel combinations, and produce novel thoughts. An airline reservation system can learn about your specific travel plans, but it couldn't learn a new fact about air travel through verbal communication.

If the world could be hand-coded into a representation like that of an airline reservation system, that could serve as the semantic basis for a communication system. The problem is that hand-coding the world is enormously difficult, so the technique of most NLP programs is to attempt to develop a system that will learn on its own from existing data. So far, those attempts have not resulted in structured knowledge of the sort one can create in a hand-designed system. We are not arguing that such systems cannot be constructed, and if they could be, they would serve as potential models of discourse. However, the practical problems of constructing them have yet to be solved.

The second type of dialogue system is more akin to large-scale language models; they are typically broader in scope and trained on very large corpora of text-based dialogue \citep{Serban2016a,Sordoni2015}. Although their language skills are much improved, these so-called ``chit chat'' systems are characterized by their undirected and reactionary nature. These systems are trained to react to user utterances in statistically appropriate ways, rather than formulating words based on internal desires, goals, and plans. If these systems have analogs to these internal states, it's only in the most implicit sense. Their lack of grounding has been one factor critics have cited for the unfortunate tendency of some models to echo offensive language, as the models do not have representations of what their speech actually refers to, nor how listeners may respond to it \citep{Bender2021}. Standard models do not have inductive biases that encourage the formation and use of goals and desires, which will be necessary for future models of conversation to be successful as accounts of psychological semantics.

There are ongoing efforts focused on addressing some of these shortcomings. Models can be conditioned on ``persona'' embeddings to encourage consistency in personality and goals, although again in a highly implicit sense \citep{Li2016b,Zhang2018}. Additionally, these systems are rarely grounded in the ways discussed in the first desideratum, but new efforts have focused on more grounded forms of dialogue through curriculum-driven language learning \citep{Mikolov2015}, text-based adventure games \citep{Urbanek2020} and discussions of natural images \citep{Shuster2018}. Indeed these directions may expand machine capabilities, but as it stands, humans are unique in choosing words on the basis of genuine ideas, desires, and plans. This ability is thanks to the deep links between their internal states and the meaning of words. If one wishes to develop a NLP system based purely on textual input, it will be difficult to learn internal representations that can direct word use.

\subsection{Word representations should support responding to instructions appropriately.}

The third desideratum concerns turning words into actions. None of the models discussed so far can respond to instructions, beyond simply generating more text. We do not fault them for this; our discussions have focused on models with sophisticated word representations rather than models that take actions in the world. Nevertheless, just as understanding words as people do requires more than hooking up a language model to a camera or a computer vision system, simply attaching a robotic (or virtual) body to an NLP system and teaching it some instructions will not suffice. Suppose we made such a language-action hybrid system and asked it to follow a new instruction, ``Pick up the knife.'' The word embedding for \emph{knife} (or the sentence representation as a whole) would need to convey a lot of information about the kind of object a knife is, in order for motor commands to properly operate on it. Although the embedding's nearest neighbors indicate related objects, parts, and functions (\emph{dagger, blade, sword,} and \emph{slicing} according to BERT), the representation for \emph{knife} must contain enough structured information so that the listener picks up the knife rather than the spoon. Although a model need not know everything about knives, it will need to know their basic functions, parts, and uses in order to understand even the simplest conversations involved them. However, after telling GPT-2 to ``Pick up the knife,'' it offered us this questionable continuation of the text: ``If the blade is still on, place it in your pocket.'' If a model has encountered the word \emph{knife} only in some sentence contexts, like ``put down the knife,'' ``dropped the knife,'' or ``use the knife,'' it must be able to understand and use the word correctly in new sentential contexts, like ``point the knife'' or ``broke the knife.'' This may be more difficult than it appears (see discussion below of \citeauthor{Ruis2020}, \citeyear{Ruis2020}).

Obviously, one should not expect text-based models to understand and perform actions. Text-based models need to be combined with other classes of models, as recent work focused on multi-modal models for instruction following does \citep{Hill2017,Hill2020,Ruis2020,Hill2020b}. These systems typically follow the architectural blueprint discussed earlier (Figure \ref{fig_arch1}), using a visual encoder to process visual input (a 3D room or 2D grid with objects in it) and a language encoder to process instructions (``Walk to the red circle'' or ``Put the picture frame on the bed''). As output, the network produces actions aimed at successfully completing the target instruction. After extensive training---often millions or billions of steps---these models typically understand enough about their subset of language, action space, and environment to successfully complete basic instructions. But do these models understand the words that they act upon? Are their representations of the words flexible enough that they can follow instructions with novel combinations that did not exist in the training set, as in our \emph{knife} example?

In many cases, instruction-following models can make meaningful generalizations. After learning to find various types of objects, models can often generalize to novel object and color pairs, successfully ``finding the fork'' when the fork is presented in a new color (\citeauthor{Hill2017}, \citeyear{Hill2017}; but see also \citeauthor{Ruis2020}, \citeyear{Ruis2020}). After learning how to ``find'' or ``lift'' across many scenarios consisting of 30 different objects, models can generalize successfully to ``lifting'' objects that they only had experience with ``finding'' (or vice versa) \citep{Hill2020}. After learning that a certain class of heavy objects require more actions to successfully ``push'' them, current models generalize that these same objects require more actions when ``pulling,'' given experience pulling other classes of heavy objects \citep{Ruis2020}.

The same class of models, however, struggles with many other aspects of instruction following. For instance, models struggle to learn an abstract, composable meaning for the word \emph{not}, failing to ``Find \emph{not} the ball'' even after learning about \emph{not} in many training episodes with other types of objects. After learning \emph{not} with respect to dozens of object types, generalization to new objects is below 50\% correct in a 2D grid world (20 training object types), and somewhere between 60\% and 80\% in a richer 3D environment \citep{Hill2020}. The model's semantic representation of \emph{not} is insufficiently abstract and too grounded in the particulars of its training experience.

\begin{figure}[t]
	\centering
	\includegraphics[width=.9\linewidth]{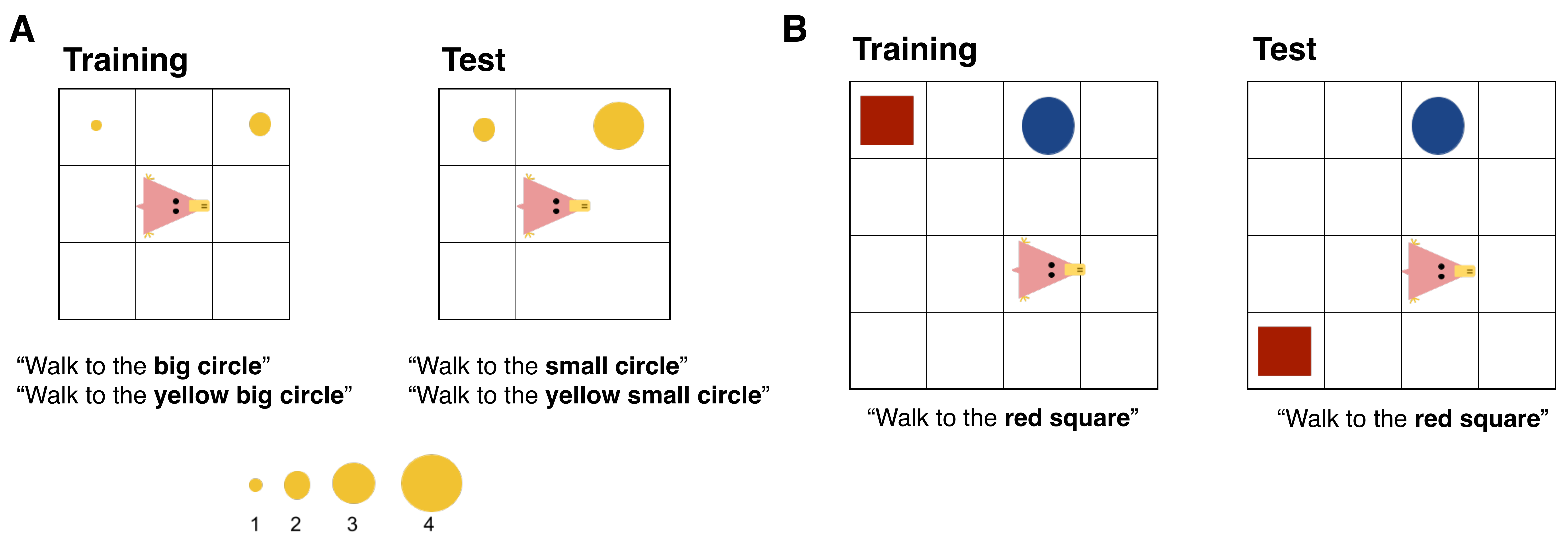}
	\caption{Instruction following in the gSCAN benchmark. (A) Generalizing from calling an object ``big'' to calling it ``small.'' (B) Generalizing to walking to targets in the south west, after learning to find targets in all other directions. Modified with permission from \citet{Ruis2020}.}
    \label{fig_gSCAN}
\end{figure}

Standard models often fail to acquire abstract, composable meanings of other types of words, such as actions. The learned concept for \emph{move} is too tied to the training experience, e.g., ``Move to the red square'' as implemented in a simple grid world (Figure \ref{fig_gSCAN}B). In the recent Grounded SCAN benchmark, \citeauthor{Ruis2020} showed that you can train an agent to expertly \emph{move} to targets positioned due south, due west, or anywhere except to the southwest of the agent's current position (which is held out for testing). At test, the agent fails catastrophically when moving to a target to its southwest (0\% correct). The agent often moves the appropriate number of steps west, or the appropriate number of steps south, but cannot seem to do both together to actually reach the target. Based on its attention maps, it seems to understand the target location but not how to ``move'' there.

The same model fails to learn abstract meanings for relational words, including \emph{small} and \emph{large,} that depend on the context, i.e., the sizes of other objects in the display \citep[Figure \ref{fig_gSCAN}A;][]{Ruis2020}. Similarly, agents struggle to learn abstract meanings for adverbs such as \emph{spinning} or \emph{zigzagging,} when tasked with instructions such as ``Move to the red square while \emph{spinning}.'' Although agents are great at spinning and zigzagging for learned instructions, they are abysmal when required to do these things in untrained combinations or novel scenarios.

In sum, instruction-following models have a long way to go before understanding words as a person does. Current neural network models rely too much on pattern recognition, learning to identify high-value states or mapping chunks of instruction to chunks of action, without sufficiently grappling with more abstract forms of meaning \citep{Lake2016}. Correctly following instructions requires forming a representation of the speaker's desired goal and connecting this to the physical world, which may require more abstract representations than current models have. We are not claiming that models will never be able to reach these goals. Recent work has begun exploring ways of utilizing pre-trained language models for instruction following tasks, enriching the underlying word representations and facilitating transfer to novel synonyms and more natural forms of instruction \citep{Hill2020b}. Other work has explored meta-learning as means of encouraging generalization to new situations and novel combinations of instructions \citep{LakeMeta2019}. However, the improvements via these approaches have been modest compared to the flexibility people demonstrate in interpreting familiar words in new ways when responding to instructions. There is still much work to do before they can be considered potential theories of psychological semantics.

\subsection{Word representations should support producing and understanding novel conceptual combinations.}

Language has been characterized as making infinite use of finite means, allowing familiar elements to combine productively to make new meanings. Our paper focuses on word meaning rather than language more generally, so we cannot discuss all the ways in which language is compositional \citep[see][for failures of NLP models on simple relational phrases]{Gershman2010}. Nevertheless, the compositional nature of language constrains models of word representation; any model of word meaning that does not permit compositionality would be a non-starter from the perspective of psychological semantics. To focus on semantic rather than syntactic composition, this section considers \emph{conceptual combinations}: two-word noun phrases that include adjective-noun (e.g., ``dirty bowl'') and noun-noun phrases (e.g., ``apartment dog''). Speakers produce novel compositions during conversation, and listeners can understand these compositions \citep{Wisniewski1997}. Although a processing account is required to combine noun representations, a model of psychological semantics should provide the appropriate information to allow such a process to take place. As we noted above, such combinations can require significant background knowledge to produce an appropriate interpretation \citep{Murphy1988}.

Do modern NLP systems provide a basis for understanding combinations? Within the distributional semantics tradition, there have been attempts to create vectors for phrases like ``apartment dog'' out of the vectors of the two words. These attempts have had some successes in accounting for various human judgments \citep{Fyshe2015,Gunther2016,Marelli2017,Vecchi2016}; for example, one may assess a model by computing the similarity of its interpretations of two phrases and seeing if they correlate with human judgments of similarity \citep[e.g.,][]{Mitchell2010}. However, as we pointed out in our discussion of early scaling approaches to word meaning, a model can account for the similarity of word meaning without actually providing information necessary to use the word. In this case, there is no easy way to find out how the model is interpreting ``apartment dog,'' even if we know that its interpretation is similar to that of ``tame lion,'' say. Furthermore, \citet{Yu2020} have discovered that models can do well in the similarity task even if they don't actually combine the representations of the two nouns, because similarity judgments are predicted by the similarity of the individual nouns. When that cue is removed, models' success dropped precipitously. It would be useful, then, to discover what features models attribute to a combination, especially features that are not true of the individual nouns, \emph{emergent features}. That would reveal a truly combinatorial process.

As we suggested above, it is possible that through their processing of billions of sentences, models have in some sense learned the knowledge necessary to process combinations, at least in terms of verbal relations. That is, perhaps the model has learned what goes in bowls, when they are washed, what makes things sticky, etc., through thousands of sentences that refer to such things. Thus, perhaps they can understand conceptual combinations like ``dirty bowls are sticky'' (see below) fairly well. To address this question, we first review some of the work on conceptual combination in cognitive science and NLP. Second, we compare people and modern Transformers on a set of complex concepts for which emergent features have been identified in past research.

Any theory of concepts must ultimately account for conceptual combination, and a variety of ideas have been put forward in the cognitive science literature about how to do that (\citeauthor{Murphy2002}, \citeyear{Murphy2002}, Chapter 12; \citeauthor{Wisniewski1997}, \citeyear{Wisniewski1997}). One can construct a fairly simple theory of adjective-noun composition using feature weighting and adjustment \citep{Smith1988}. For the combination \emph{purple jacket}, the head \emph{jacket} is a prototype in feature space and the adjective \emph{purple} changes a particular feature of that prototype, using a self-contained process that depends only on this pair of word representations. However, this account cannot explain noun-noun compounds, which are extremely common in English, and many modifiers don't seem to adjust the same features when combined with different heads, e.g., \emph{ocean view, ocean book, ocean wave}, etc. \citep{Murphy1988}. Furthermore, although a purple jacket is purple, an ocean book is not ``ocean.'' These complexities suggest instead that conceptual combination requires active interpretation in ways that rely on background knowledge, using knowledge to decide which features to modify and how they should be adjusted. For instance, \emph{empty stores} typically lose money, but neither \emph{stores} nor \emph{empty things} typically do so. We infer this through knowledge that stores make money by selling products, customers purchase products, empty stores have no customers, etc., again revealing the close connection between conceptual knowledge and language understanding. Although knowledge-based accounts are difficult to formalize, they are necessary to account for the sophistication of human conceptual representations and the mental chemistry by which they combine.

Research in NLP has also examined how word embeddings might be combined to form complex concepts \citep[see][for a review]{Mitchell2010}. When introducing word representations above, we discussed averaging word embeddings (or summing) as a means of composition in NLP, e.g., ``Vietnam'' + ``capital'' results in a vector similar to that of ``Hanoi'' \citep{Mikolov2013a}. We quickly dismissed averaging as a means of constructing sentence representations, due to the loss of syntactic and relational structure, but these issues are less damaging for two-word compounds. Nevertheless, many of the cases discussed above pose challenges to additive models, namely modifiers that affect different head nouns in different ways \citep[see][]{Murphy1993}. For example, an \emph{ocean view} is a view where one can see the ocean, whereas an \emph{ocean wave} is a wave of the ocean. It's difficult to see how adding the same exact \emph{ocean} vector to each head noun could produce such a wide range of semantic transformations.

More capable models use matrix-vector multiplication with adjectives as matrices and nouns as vectors \citep{Baroni2010}. Linear transformations can account for more subtle transformations, including adjectives that emphasize one property for some types of nouns and another property for other types of nouns \citep{Baroni2014a}. Still, emergent properties that require active interpretation and background knowledge remain a clear challenge for simple models of composition (e.g., ``empty stores lose money''). Instead, large-scale Transformers allow for essentially arbitrary transformations, at least in principle, and thus we compare this model class with human judgments concerning conceptual combination.

In a series of human behavioral studies, \citet{Murphy1988} studied the role of background knowledge in conceptual combination by collecting ratings of whether certain features are more typical of compounds or of their constituent parts. He reasoned that if conceptual combination is a closed operation---and the features of compounds can be computed locally from the features of its constituents---compounds should not have properties that are not found in their constituents, emergent features. He constructed 18 adjective-noun compounds paired with properties that he hypothesized would violate this constraint, in that forming the conceptual combination requires background knowledge. Human participants were asked to rate how typical a feature (e.g., ``loses money'') is of a category; some participants made these ratings with respect to the compounds (e.g., \emph{empty stores}), others with respect to the adjective (\emph{empty}), and others with respect to the noun (\emph{stores}). Murphy found that for 15 of the 18 items, the candidate property was rated as more typical of the adjective-noun compound than either the adjective or bare noun. We show these 15 items in Table \ref{table_complex_concepts}. For instance, \emph{sliced apples} are typically ``cooked in a pie,'' but neither \emph{apples} nor \emph{sliced things} are as typically associated with that property. \emph{Dirty bowls} are typically ``sticky,'' but neither \emph{bowls} nor \emph{dirty things} are typically sticky. Thus, these examples could not be accounted for by a model that draws only on the two noun representations to arrive at a phrasal interpretation.

NLP systems cannot be directly queried about how typical a property is of an object, yet as probabilistic language models, they implicitly associate objects and properties. These associations can be probed in various ways; we chose the conditional probability $P(property|object)$---as estimated by autoregressive models such as GPT-2---to be a straightforward means of extracting these associations. (BERT cannot be easily evaluated since it predicts only the probability of individual missing tokens.) To measure the association between a noun phrase and ``lose money,'' GPT-2 was queried using an association score, $P(\text{lose money.}|object)$, such that $object$ can be filled by ``Empty stores,'' ``Crowded stores,'' ``Stores,'' ``Regular stores,'' or ``Empty things.''\footnote{Periods were always included at the end of the sentences in the evaluations.} Although the raw probabilities are uninterpretable, their relative rankings are informative: How much more likely is the probe ``Empty stores lose money'' compared to ``Regular stores lose money''? The scores for different objects can be meaningfully compared because this method controls for differences in the noun phrases due to a variety of factors: frequencies, lengths, and prior probabilities. This control is possible because the left-hand-side of the conditional is the same in each case, and the different noun phrases appear only on the right-hand-side. 

Using this methodology, we evaluate GPT-2's ability to predict the 15 items from \citet{Murphy1988} that yielded emergent features. Recall that in the behavioral study, people were evaluated on the relation between the property and the complex concept ``Empty stores,'' compared to the property and the bare noun (``Stores'') and the bare adjective (operationalized as ``Empty things''). GPT-2 was evaluated on these three relations too. For additional rigor, we also evaluated the model on an alternative phrasing of the bare noun (``Regular stores'') and an inconsistent adjective-noun pair (``Crowded stores''). A prediction was considered successful (and in alignment with the human judgments) if the property was most typical of the consistent adjective-noun pair, e.g., if ``Empty stores lose money'' had the highest score.

As summarized in Table \ref{table_complex_concepts}, GPT-2 makes successful predictions for only 7 of 15 items. Notable errors include judging ``are rusty'' as more typical of ``new saws'' than ``ancient saws'' and judging ``are sticky'' as more typical of ``clean bowls'' than ``dirty bowls.'' GPT-2 also judges ``losing money'' as more typical of ``regular stores'' than ``empty stores.'' Despite obvious errors, it did get some cases right which seemingly require background knowledge, knowing that ``are cooked in a pie'' is more typical of ``sliced apples'' than ``whole apples.'' However word repetition may have helped it get two other cases right: ``Russian novels are written in Russian'' and ``Green bicycles are painted green.''

To see if the model would do better with increased context, we re-ran each case with an additional sentence before the main sentence of interest. (The sentence did not directly refer to the tested feature.) For instance, GPT-2 scored the following multi-sentence utterance, ``Stores are still where most products are purchased. Empty stores lose money.'' As before, the only phrase to appear on the left-hand-side of the conditional is ``lose money'' while everything else appears on the right-hand-side. The additional context did not seem to help, as the model largely got the same set of cases right (7 of 15; last column, Table \ref{table_complex_concepts}).

 This performance is consistent with \citeauthor{Yu2020}'s (\citeyear{Yu2020}) conclusion that many interpretations do not go far beyond the individual lexical items, even for sophisticated Transformers that, in principle, allow for very flexible types of conceptual combination. Of course, the results of \citeauthor{Yu2020} and the items from \citet{Murphy1988} should not be taken as the final word on understanding complex concepts in sophisticated language models. We suggest that other tests of combinations sample the content of the representation rather than similarity between combinations. Our test may also have been easy in some respects. Due to the massive scale of training, predictive text models like GPT-2 will be familiar with many of these complex concepts already (empty stores, sliced apples, etc.)---even if it doesn't understand them fully. In contrast, people can generate and understand genuinely new compositions, say a \emph{cactus pig} or \emph{snow soda} \citep{Wisniewski1997}. We see the challenge of understanding complex concepts, and the role of background knowledge in interpreting these compositions, as key to understanding words as people do.

\begin{table}[t]
\centering
\caption{\\Comparing humans and GPT-2 on properties of complex concepts. Most typical part shows which noun phrase (e.g., Empty stores, Crowded stores, Stores, Regular stores, or Empty things) participants \citep{Murphy1988} or models judged the indicated property to be most typical.}
\label{table_complex_concepts}
{\footnotesize
\begin{tabular}{lllll}
& & \multicolumn{3}{c}{\textbf{Most typical part}}\tabularnewline
\textbf{Combination} & \textbf{Property} & \textbf{Human judgments} & \textbf{GPT-2 (no context)} & \textbf{GPT-2 (context)}\tabularnewline
\hline
Sliced apples & are cooked in a pie. & Sliced apples & \textbf{Sliced apples} & \textbf{Sliced apples} \tabularnewline
Casual shirts & are pulled over your head. & Casual shirts &
Formal shirts & Casual things \tabularnewline
Small couches & seat only 2 people. & Small couches & \textbf{Small couches} & \textbf{Small couches} \tabularnewline
Uncaged canaries & live in South America. & Uncaged canaries & Canaries & Canaries \tabularnewline
Round tables & are used at conferences. & Round tables & \textbf{Round tables} & \textbf{Round tables} \tabularnewline
Standing ostriches & are calm. & Standing ostriches & Standing things & Ostriches \tabularnewline
Unshelled peas & are long. & Unshelled peas & Unshelled things & Regular peas \tabularnewline
Yellow jackets & are worn by fishermen. & Yellow jackets &
Regular jackets & \textbf{Yellow jackets} \tabularnewline
Green bicycles & are painted green. & Green bicycles & \textbf{Green bicycles} & \textbf{Green bicycles} \tabularnewline
Overturned chairs & are on a table. & Overturned chairs & \textbf{Overturned chairs} & \textbf{Overturned chairs} \tabularnewline
Short pants & expose knees. & Short pants & \textbf{Short pants} & \textbf{Short pants} \tabularnewline
Ancient saws & are rusty. & Ancient saws & New saws & Saws \tabularnewline
Russian novels & are written in Russian. & Russian novels &
\textbf{Russian novels} & Russian things \tabularnewline
Empty stores & lose money & Empty stores & Regular stores & Regular stores \tabularnewline
Dirty bowls & are sticky. & Dirty bowls & Dirty things & Dirty things \tabularnewline
\end{tabular}

\vspace{1ex}
{\flushleft Note: Bold indicates a match between human and model judgments. \par}
}
\end{table}

\subsection{Word representations should support changing one's beliefs about the world based on linguistic input.}

A theme so far has been how knowledge influences one’s understanding of words and sentences. However, the influence also runs in the other direction: Language is a source of knowledge and, with perception, one of the main inputs into our beliefs about the world. If you hear in elementary school, “Sharks are fish, but dolphins are mammals,” this may change your understanding of all the concepts mentioned, permanently. And then you may act on this knowledge, both in terms of verbal output and actions. The same is true for more specific and mundane sentences, such as “I left the car on the street.” Thus it is important that linguistic representations be interfaceable with one’s knowledge of the world.

Whether one thinks this is possible or impossible for current NLP models depends on one’s attitudes towards the text-based training of large-scale language models like BERT and GPT. Given that the representations formed by such systems are based on texts and relations among word parts, it is arguable that its representations do not constitute knowledge of the world, and therefore, its representations of sentences cannot add to a knowledge store: All they can do is tell us about complex relations among words. We believe that there is much to be said for this perspective \citep[and see][]{Bender2020}. However, it is possible to take a more generous view of the models and ask whether they could serve as part of a knowledge base, that when linked to perceptual or motor modules, would give them substantive content.

Therefore, we consider here to what degree NLP models might represent a knowledge base that can take input from language and give reliable information. Predictive models such as BERT, GPT-2, CBOW, etc. are not incentivized to learn about the world per se, only to learn about how other words in the context predict a target word. However, one could conceive of this information as approximating knowledge of the world. If many sentences describe the actual world, then text predictions could be used to learn those descriptions, which constitute its beliefs about the world (in text). This in turn leads us to ask whether such beliefs are reasonable and also whether they can be readily added to and changed through linguistic input.

Changes in a model's lexical representations should be reflected, in some way, through changes in its knowledge about the world. For instance, a language model is tasked with predicting a missing word in a novel sentence, ``Sharks are fish but dolphins are \mask''. A model that knows little about dolphins might predict \emph{fish} with high probability, and then receive feedback that the correct answer was actually \emph{mammals}. Through backpropagation, the model makes small adjustments to increase the probability of outputting \emph{mammals} when this sentence is encountered again. These adjustments will change the input layer's word embeddings (for \emph{fish, sharks, dolphins,} etc.), the output layer's word representations (\emph{mammals,} \emph{fish,} etc.), and many other internal parameters. After sufficient presentations of the sentence, the network will learn to produce the right answer, hopefully in a way that generalizes to related sentences. But is such learning sufficient to develop correct beliefs about dolphins, mammals, and their properties?

It's not entirely clear how to evaluate the beliefs of a language model. Although other methods are possible, we tested models as closely as possible to the way they were trained, by providing sentence fragments with missing words to be completed. Our informal tests focused on taxonomic category relations such as category membership that have been central to research in semantic memory in people \citep[e.g.,][]{Rips1973}. Some predictions will certainly be correct; when probed with ``Sharks are fish but dolphins are \mask'',\footnote{As in previous tests, periods were used at the end of the sentences, and BERT's special tokens were added at the beginning and end of the sentences.} a trained BERT predicts that \emph{mammals} (probability 0.027) is more likely than \emph{birds} (0.013) and \emph{fish} (0.010). However, BERT's predictions are sensitive to small changes in wording---a property of current NLP systems that has been observed elsewhere \citep{Jia2017,Marcus2019}. When probed with ``Sharks are fish \emph{and} dolphins are \mask'' (swapping \emph{but} with \emph{and}), BERT now predicts that \emph{birds} (0.14) is more likely than \emph{mammals} (0.031) or \emph{fish} (0.025). Similarly when probed with ``Sharks are fish \emph{while} dolphins are \mask'', BERT predicts \emph{birds} again. If probed more directly as ``Dolphins are \mask'', BERT now properly predicts \emph{mammals} (0.0024) over \emph{fish} (0.0014) and \emph{birds} (0.00031). But when asked the same question about the singular noun, ``A dolphin is a \mask,'' it now predicts \emph{fish}. Other work shows that BERT is especially poor at understanding the negated versions of such probes, e.g., ``A shark is NOT a \mask'' \citep{Ettinger2019}. Examining the word embeddings doesn't clear up story. For BERT, \emph{dolphin} is more similar to \emph{mammal} (the cosine is 0.40) than it is to \emph{fish} (0.32) and \emph{bird} (0.33). However, \emph{dolphin} is more similar to \emph{fishes} (0.45) than it is similar to either \emph{mammal} or \emph{fish} (or \emph{mammals} or \emph{birds}). It seems that BERT has some ideas about what dolphins are, but it is too tied to specific wording to be credited with general knowledge.

In a more systematic comparison, we evaluated knowledge of 31 animals in BERT and GPT-2 using the same framing as before, ``A squirrel is a \mask'' and ``Squirrels are \mask''.\footnote{The list of animals in \citet{Kemp2008} was used  after removing the two ``reptiles'', since this word isn't a single token in BERT. We didn't alternate ``a'' vs. ``an'' in the questions, which could help reveal the answer.} We considered four possible superordinate categories as answers: \emph{bird/birds}, \emph{insect/insects}, \emph{mammal/mammals}, and \emph{fish/fish}, using the singular or plural form of the category depending on the question. BERT predicts the right answer (i.e., highest probability for the correct superordinate of the four) for only 54.8\% of the questions in singular form, and for  77.4\% of the questions in plural form. (Note that chance performance would be 25\% correct.) Strangely, the two forms of the same question yielded inconsistent answers more often (51.6\%) than consistent answers. It predicts  \emph{squirrels} and \emph{horses} are \emph{mammals}, but \emph{a squirrel} or \emph{a horse} is \emph{a bird}. \emph{Whales} are \emph{mammals}, but \emph{a whale} is \emph{a fish}. \emph{Butterflies} are \emph{birds}, and so is \emph{a butterfly}. GPT-2 fared better with 83.9\% accuracy on singular forms and 77.4\% for plural forms. Still, there was a striking inconsistency between the two ways of asking the same question, with 35.5\% mismatches in the answers.

Finally, we evaluated knowledge of animal parts using a similar methodology, ``A wolf has \mask'' or ``Wolves have \mask''. We considered ``legs,'' ``fins,'' and ``wings'' as possible answers, evaluating the subset of the above animal categories for which these answers are mutually exclusive (19 categories total). Surprisingly, for the singular nouns, BERT chose ``wings'' as the best answer in every case (0\% correct; no birds or flying insects were analyzed since they have both legs and wings). For the plural forms, accuracy was somewhat higher (42.1\%). GPT-2 fared a bit better: accuracy was 47.3\% for the singular nouns and 36.8\% for the plural nouns.

Taken together, it's unclear if current language models hold any genuine and consistent beliefs about basic taxonomic and part-whole relations. This uncertainty persists despite training on billions of words that include, presumably, the entire Wikipedia entries for \emph{dolphins}, \emph{mammals}, \emph{fish}, etc. This muddiness is the hallmark of a primarily pattern-recognition driven learning process. During training, BERT hones its abilities at predicting missing words and the order of sentences, acquiring some inkling of how the word \emph{dolphins} is related to the words \emph{mammals},  \emph{fish}, and \emph{flippers}, but nothing seemingly explicit or belief-like. As a result, it is ineffective at changing its beliefs or building a coherent world model based on systems of beliefs \citep{Lake2016}.

The challenges of representing and changing beliefs extend far beyond just taxonomic categories, and text generation provides another window into what, if anything, language models believe. Autoregressive models such as GPT-2 can generate impressive passages of text, although these models frequently contradict themselves. In a highlighted demonstration of GPT-2's text generation capabilities \citep{Radford2018a}, the model was tasked with reading a fanciful prompt concerning talking unicorns and producing a reasonable continuation:
\begin{quote}
\emph{Prompt}: In a shocking finding, scientist discovered a herd of unicorns living in a remote, previously unexplored valley, in the Andes Mountains. Even more surprising to the researchers was the fact that the unicorns spoke perfect English.

\emph{GPT-2's continuation}: The scientist named the population, after their distinctive horn, Ovid's Unicorn. These four-horned, silver-white unicorns were previously unknown to science. Now, after almost two centuries, the mystery of what sparked this odd phenomenon is finally solved\ldots
\end{quote}
GPT-2 impressively generates several more paragraphs of seemingly natural text \citep[][pg. 13]{Radford2018a}. The model understood enough about this bizarre scenario---unlikely to be in its training corpus---to write fluently about it. Nevertheless, a closer look is revealing about the structure of GPT-2's beliefs, in this particular case, as related to unicorns. In the generated passage, GPT-2 contradicts itself almost immediately, writing about ``these \emph{four-horned,} silver-white unicorns.'' The model, evidently, doesn't understand that unicorns must have one horn. Also, in two adjacent sentences, it suggests that unicorns have one horn (``their distinctive horn'') while simultaneously having multiple horns (``four-horned''). Even if you could help correct GPT-2, as you would a person, by specifying that ``unicorns must have just one horn,'' it's doubtful that GPT-2 would get the message, since any one sentence would have a trivial effect on its representations. Indeed, it seems hard to believe that similar sentences were not already in its learning corpus, given that it knows something about unicorns. It's also not clear how one would open up the model and add/correct this fact, given that GPT-2 uses 1.5 billion learnable parameters to make predictions.

Language models can answer some difficult factoid-style questions, although they are hardly a reliable source. Using a corpus called Natural Questions \citep{Kwiatkowski2019}, GPT-2 was evaluated on factoids such as ``Where is the bowling hall of fame located?'' or ``How many episodes in season 2 breaking bad?'' GPT-2 answers 4.1\% of these somewhat obscure questions correctly \citep{Radford2018a}; performance is higher in the larger-scale GPT-3 \citep[15-30\%; ][]{Brown2020} and can also be substantially improved when combined with techniques from information retrieval \citep{Alberti2019}, e.g., providing BERT with all of Wikipedia and allowing it to answer by highlighting passages. Other studies \citep{Petroni2019} have found that BERT makes reasonable predictions for more commonplace questions from the ConceptNet knowledge base, including ``Ravens can \mask'' (prediction: \emph{fly}) and ``You are likely to find a overflow in a \mask'' (prediction: \emph{drain}), although these predictions are brittle in all the ways outlined above. This suggests that current methods of extracting information from text corpora have not yet formed knowledge bases that would be sufficient for conceptual combination and language understanding more generally, although techniques continue to improve.

Computational theories of semantic memory, from the very beginning, have recognized the need to specify the relations between words (or concepts) in order to provide coherent, accurate representations. That is, models must know that a dolphin ISA mammal and CAN swim, or else it will not be able to correctly draw inferences. Unlabeled links from \emph{dolphin} to \emph{mammal} and \emph{swims} are not sufficient \citep[e.g.,][]{Brachman1979,Collins1968}. Word representations need to represent which words are properties, parts, synonyms, objects acted on, or things that simply tend to co-occur with the word's referent. Earlier neural network models attempted to capture this type of semantic knowledge and in particular, the developmental process of acquiring semantic knowledge \citep{Rogers2004}. The limitation of the approach was not theoretical, but practical: The modelers hand-coded the features and category names and their relations. Thus, the model was told that a canary is colored yellow, not just that canary and yellow go together. The argument for this is that human children get these relations through both perception (yellow is a color, which the canary-learner already knows) and language (parental input). Thus, specifying these relations is not a \emph{deus ex machina} merely designed to make the model work but an attempt to provide the information that perception would for a person. The NLP models we discuss do not attempt to learn explicit relations; they rely purely on text predictability to provide all the information. 

Two things need to be done in order to scale up the \citet{Rogers2004} approach. First, perception must ground relations between properties and objects that are only implicit in language, using techniques in computer vision (Desideratum 1). Second, to avoid hand-coding, a neural network should be developed that can extract relations between words (ISA, CAN, etc.) from text even if it does not explicitly include labeled relations. As discussed in Section \ref{sec_NLP_models}, structured distributional models aim to do exactly this by looking for specific word patterns \citep{Baroni2010b,Baroni2014a}. The system could also use a knowledge base to encourage explicitness and consistency in its belief system, relating to current efforts that combine neural networks with knowledge bases \citep{Bosselut2020}. To be fully successful, however, such a hybrid model would need to be able to use words to change its beliefs (such as ``unicorns have only one horn''), as opposed to merely accessing fixed beliefs, further highlighting the complexity of psychological semantics and the abilities it supports.

\subsection{Summary of Desiderata} \label{sec_summary_desiderata}

Our critique is not that NLP researchers have failed to provide us with robots that converse about the world and follow our orders. From the perspective of psychological semantics, our critique is that the current word representations are too strongly linked to complex text-based patterns, and too weakly linked to world knowledge. Multi-modal models can enrich these word representations by grounding them in vision and action, yet these word representations are currently too limited by the particulars of their previous experiences. More abstract semantic representations that connect language to knowledge of the world are needed to capture the way that language leads to action and knowledge in people.

\section{Successes in NLP}

This raises the question of why deep learning for NLP works so well on many important problems. A full accounting of the remarkable successes of deep learning is beyond the scope of this article; they have also been discussed and analyzed at length in many places \citep[e.g.,][]{Lecun2015,Schmidhuber2015}. The reemergence of neural networks in the last decade was catalyzed by successes on quintessential pattern recognition problems, particularly object recognition \citep{Krizhevsky2012} and speech recognition \citep{Graves2013a,Hannun2014}, by learning features from raw data that were previously hand-designed. It's natural to think that this approach would make advances in NLP as well, especially when combined with innovations in architecture \citep{Hochreiter1997,Vaswani2017}, large datasets, and enormous computing resources. These successes are amplified by taking pre-trained word embeddings, or whole language models, and fine tuning them to perform particular tasks, as is the typical current approach for tackling NLP benchmarks \citep{Wang2019a,Devlin2019}. Modern NLP systems, following this model, know a surprising amount about syntax, semantics, and even enough to answer some basic questions about the world \citep{Rogers2020}. As discussed in Section \ref{sec_similarity}, word embeddings know enough about the relationships between words to predict human similarity judgments and related tasks. The result of this is that the models can do well when they are given words as inputs and words as output, learning the right sorts of associations and patterns through massive amounts of training. The surprising bits of knowledge that models learn presumably come from this process.

For example, consider the world capitals that \citet{Mikolov2013a} tested models on. The models don't have a list of countries and their capitals; they don't know what a country or a capital is, which would be required to use the word correctly in conversation or writing. However, it seems likely that the words \emph{Paris, capital,} and \emph{France} co-occur fairly often and more often than \emph{Lyon, capital,} and \emph{France} or \emph{Paris, capital,} and \emph{Argentina} do. Verbs and their past tenses or inflections also co-occur, as the same action is talked about in infinitives, and in different temporal contexts. For example, it would not be surprising if a text about surfing contained sentences like, ``Stephanie wanted to surf...,'' ``She had surfed before...,'' and ``Surfing is dangerous when...'' The things that one says about surfing one might say in talking about past, present, and future contexts, so that the representations of different forms of the verb could become similar. The same is true for nouns and their plurals. These models don't know what a plural is, but they can learn that \emph{knife} and \emph{knives} occur in the same kinds of passages, and so they are assigned similar representations. To a limited extent, language models can even track long distance syntactic dependencies, knowing it's proper to say ``the knives in the drawer \emph{cut}'' rather than ``the knives in the drawer \emph{cuts}'' \citep{Linzen2016}. However, based on text distributions alone, the models don't necessarily learn that \emph{knife} refers to one thing and \emph{knives} to multiple things. The models don't know that there are things. Without a representation of what a knife actually is, it cannot form a semantic representation of the sort that people have. 

There is a lot of text in the world, more than some of us realized before we began to read about corpora of 630 billion words. Finding relations among textual entities can therefore be extremely useful. Furthermore, when people read the outputs of such models, they can fill in the semantic gaps themselves to understand what the model has found. We argued at the beginning that words ultimately gain their meaningfulness by connecting to the world. Humans can provide that connection, when the model produces textual output and the human connects the text to the actual world. If a particular word has the following LSA neighbors, \emph{sculptured, sculptor, sculpture, Acropolis, colonnade, Athena, Parthenon,} and \emph{gymnasiums}, readers can readily figure out that this word must have something to do with artwork common in ancient Greece, found in temples, etc. But this is the human interpretation of the word, not something the model has told us. When the human and the model work together, they may be able to interpret unknown words in a way that goes beyond the model's own performance. We suspect that this is part of the reason why researchers have taken these models (especially the early count-based ones) seriously as theories of word meaning. When the researchers (or anyone) read the list of nearest neighbors, \emph{they} are identifying the links to the target word by using their own knowledge to fill in gaps and infer the underlying meaning \citep{Bender2021}. They may know, for example, that the Parthenon is a temple dedicated to Athena found on the Acropolis in Athens with statues and altars dedicated to her. Thus, they can interpret how all these words are connected to the target word \emph{statue}. Models, lacking that knowledge and inference ability, cannot do so. Of course, we often operate on the basis of charity in interpreting the utterances of human speakers, giving them credit for understanding things when they may not. However, we also know that humans can generally use language interactively in the world in the way that text-based models cannot. If an NLP system with perception and motor control can respond to instructions and give descriptions, it will have earned an assumption of charity in assuming that its utterances reflect knowledge.

Although NLP models may not understand words the way that people do, they nonetheless can be very useful for practical and theoretical uses. A number of models take the output of initial NLP prediction models as their starting point for the representation of word meanings. They can then use them to create a specific task-based system (e.g., information retrieval or restaurant reservations) or for more theoretical purposes. For example, \citet{Lu2019a} used the word vectors of Word2vec as the starting point for their model of verbal analogies. They created a Bayesian model that identified the semantic features most useful for identifying different relations used in the analogy task. The model was able to identify the most relevant relation between \textit{weaver} and \textit{cloth}, for example, and then identify that \textit{baker} and \textit{bread} have the same relation. So long as such models stay within the realm of verbal tasks, they may find that NLP representations work well. The  \citeauthor{Lu2019a} model was better at solving analogies with some relations (like the profession-product relation just shown) than others (e.g., antonyms). Whether this is a consequence of the analogy model or the Word2vec representations requires further exploration.

We now see the research frontier shifting from problems of pattern recognition to problems in reasoning, compositionality, concept learning, multi-modal learning, etc. Further progress will come from improvements in the training data, but this is unlikely to be enough. A new generation of NLP systems, developed with the five desiderata in mind, would look quite different from today's systems, while also building on their successes. Innovations will be needed to achieve more realistic representations of meaning; we pointed to advances in neuro-symbolic modeling and grounded language learning as important developing areas. Additional attention will be needed on incorporating background knowledge and encouraging abstraction, so that the representations can be accessed by goals and beliefs. We hope that our five desiderata help to pose new challenges and stimulate new research on more psychologically realistic models of semantics.

One might reverse the question asked in this section (and a reviewer has done so), to ask what kind of performance would convince us that a model's semantic representations have psychological plausibility. There is clearly no firm criterion for making such a judgment. A familiar saying is that all models are wrong but some are useful, and that applies here. No computational model will provide a full account of psychological semantics in the foreseeable future. The question is whether the model's processes are plausibly similar to those of humans, possibly giving insight into human psychology. When it makes mistakes, does it make the kinds of mistakes people do? Are the things the model finds difficult the things that people do as well? It is often very revealing what models cannot do, as this suggests that there is some limitation in their input or processing, leading to psychological hypotheses about what people must be doing. Our concern in the present article is not so much model errors as the problem of omitting whole domains of human language use (e.g., labeling objects and events in the world) and differences between many models' input and that of humans'. The errors are signs of those differences, as, indeed, are cases in which models do better than humans.

\subsection{GPT-3 and Scaling Up}
We analyzed a number of NLP systems throughout this article (LSA, CBOW, BERT, GPT-2, a caption generation system, etc.), with the largest being GPT-2. Recently, its successor, GPT-3, was published by the same group at OpenAI \citep{Brown2020}. In terms of architecture and training procedure, little has changed; GPT-3 is a large-scale autoregressive Transformer with the same architecture as GPT-2. In terms of scale, GPT-3 is a marvel of engineering that is strikingly larger than GPT-2. GPT-3 has 175 billion parameters, compared to GPT-2's 1.5 billion, and was trained on large swaths of the internet for a total of about 500 billion tokens---25 times more data than GPT-2. We argued that training with more data would not itself lead to a model of psychological semantics, and thus GPT-3 conveniently offers a case study in scaling up. The model is new and much about it is still unknown; we did not analyze GPT-3 directly in our own tests as it was unavailable at the time of writing. Nevertheless, we offer some observations of what GPT-3 accomplishes and what it doesn't.

GPT-3 is a strong few-shot learner. As with GPT-2, the authors do not fine-tune the model for specific tasks; it is trained solely on predicting the next word in a sequence. GPT-3 can perform many different tasks, however, through different text-based prompts that preface the relevant query. If provided with a few examples of question answering, grammar correction, or numerical addition, it often continues the task in response to new queries. In some cases, it can handle novel tasks that are unlikely to exist in the training corpus. Its flexibility to reuse its representations for new tasks without having to re-train \citep{Lake2016} resembles the flexibility of human semantic representations. (However, in most cases the model's performance does not reach the level of previous models that have been specifically trained on just one task; \citeauthor{Brown2020}, \citeyear{Brown2020}).

In other ways, GPT-3 is no closer than GPT-2 to meeting the five desiderata for a model of psychological semantics. (Admirably, the GPT-3 article has a thorough Limitations section that we draw from here.) First, GPT-3 is trained from text alone; thus, it is limited in all the ways that all ungrounded representations of words are (Desiderata 1 and 3). Second, GPT-3 aims to predict the next word in a sequence, no matter the task or the context; instead,  humans produce words to express internal states such as goals, desires, etc. (Desideratum 2). The GPT-3 authors mention this limitation: ``useful language systems (for example virtual assistants) might be better thought of as taking goal-directed actions rather than just making prediction'' \citep[][pg. 33]{Brown2020}. Third, we don't know yet how GPT-3 performs on tests of complex concepts (Desideratum 4), although we wouldn't be surprised to see GPT-3 outperform GPT-2. GPT-3 has far more training data, and all of the complex concepts in \citet{Murphy1988} are likely covered during training. Harder tests with genuinely novel compositions would pose greater challenges.

Fourth, GPT-3 has no new mechanisms for connecting word representations to beliefs, or changing its beliefs based on linguistic input (Desideratum 5). In fact, the larger-scale corpus---combined with weaker curation and filtering compared to GPT-2---could weaken the firmness of any proto-beliefs the model does has, as there is likely more contradictory text in the training data now for a given fact. The authors report that GPT-3 frequently contradicts itself during text generation, a problem also present in GPT-2, as we discussed in Section \ref{sec_main_desiderata}. This is evident in the two generated news articles provided in the GPT-3 paper (p. 27). In one article, it's hard to understand the core proposition the article is supposed to convey (Did Joaquin Phoenix pledge to wear a tux to the Oscars, or not?). In the other, which human judges thought was the most human-like sample, GPT-3 describes a split in the United Methodist Church. GPT-3 writes that ``the new split will be the second in the church's history. The first occurred in 1968..." But three sentences later, it goes on to describe a third split in the church, ``In 2016, the denomination was split over ordination of transgender clergy...'', although perhaps it means an ``intellectual'' rather than ``physical'' split (if it knows the difference). Last, there is no evidence yet that GPT-3 has learned semantic representations that better capture abstract meaning, in the way needed for human-level instruction following (Desideratum 3).

All that said, some readers and discussants have suggested to us that GPT-3's performance is simply too good not to be ``real'' in some way. There are examples posted in social media of extended conversations with the model, as well as story completions it has written. This has provoked reactions of the sort, ``Yes, what you say about the lack of grounding and the like is true, but still, it must mean something that the model is so good.'' The model's performance in these cases truly is impressive, and we are not attempting to diminish it. However, let's consider some of these examples, like story completion or conversation. These tasks are all cases in which some text is input to the model, and it responds with text in a surprisingly fluent and appropriate way. Because of its greater computational power, which allowed the model to pay attention to much more context during learning, it is very good at generating grammatical sentences and even multi-sentence sequences and conversational interactions that are coherent. Of course, not all are, as the model can sometimes go off on a string of associations that leads far away from the original topic and does not make obvious sense. However, the fact that the model makes a coherent story, say, does not mean that it has formed a semantic representation of the beginning and constructed a sensible continuation; the model is able to provide the kinds of sentences that follow that kind of beginning in its training experience. \citet{Brown2020} report that the model does not in fact do particularly well in reading comprehension measures (Table 3.7), even relative to other models (which generally have been specifically tuned for such tasks). (For example, in the RACE dataset based on a test taken by Chinese students to evaluate their English comprehension, the best NLP model scores about 90\% and 93\% on the two versions, but GPT-3 scores 47\% and 58\%.)  That suggests to us that the model may not completely ``understand'' a passage and yet be able to write a continuation of it that fits in well with the existing text. This is not as implausible as it might seem, as there are also humans who can carry on reasonable conversations about something, and later it is discovered that they have almost no understanding of the topic at all.

Relatedly, although GPT-3 exhibits strong performance on a wide range of tasks, it performs poorly on several adversarially generated benchmarks that pit surface-level patterns against the underlying semantic content \citep{Sakaguchi2019,Nie2020}. These types of tasks pose particular challenges to pattern recognition systems, however powerful and impressive they are. The GPT-3 authors themselves note that there are limits to scaling up, writing that a fundamental limit to ``scaling up any [language] model, whether autoregressive or bidirectional – is that it may eventually run into (or could already be running into) the limits of the pretraining objective'' (pg. 33). We hope the five desiderata provide additional guidance in how to venture beyond pattern recognition and secure a more conceptual foundation for word meaning, leading to more powerful and more psychologically plausible models.

\subsection{Past Critiques Within Psychology} \label{sec:past-critiques}

The shortcomings of purely text-based models have not gone unnoticed in the psychological literature. In particular, the need for words to make connections with the world have been pointed out and debated. However, those discussions have tended to take a different focus from ours. First, the text-based approaches (primarily HAL and LSA) have often been contrasted with \emph{embodiment} theories, which rely on \emph{perceptual symbol systems} accounts of representation and meaning \citep[e.g.,][]{Andrews2014,Louwerse2007}. In this theory, symbols are not only linked to perception but in fact are perceptual or motor simulations, which capture the essence of a concept \citep[see][for a review]{Barsalou1999}. In addition to perceptual symbols, emotion is also often mentioned as a potential feature of semantic representations \citep{Deyne2018} that might not be captured by distributional models.

A second difference from our approach is that these discussions often take a purely empirical tack. They use a model to attempt to explain some human data and then ask if the fit could be improved by the addition of perceptual information (etc.)---or the reverse, seeing if a model improves the fit of human data. For example, \citet{Deyne2018} investigated whether affective and featural information improved the modeling of human similarity judgments beyond text-based word embeddings. \citet{Louwerse2007} developed a new measure of sentence coherence to see whether LSA could explain apparent embodiment effects found by \citet{Glenberg2000}. \citet[][p. 66]{Mandera2017a} found that adding text-based model data improved the prediction of priming results over human-generated feature lists.

These studies can provide insight into what information is contained in different models and measures, as well as which information is most relevant to a given task. However, they do not address the basic problem faced by pattern-based NLP models, which is how they accomplish the main goals of a theory of semantics, summarized in Table \ref{table_desiderata}. Their evaluation data are generally relatedness measures of word pairs (or pairs of sentences)---that is, they remain primarily within the realm of words. As we reviewed above, describing a situation or scene does not have words as the input, so there must be a way to generate a sentence from perception and knowledge. And carrying out an instruction or changing one's knowledge about the world cannot be done by activating words whose meanings are only other related words. By focusing on text-based tasks only, empirical tests cannot discover the main shortcomings of text-based approaches as theories of meaning, how language is used to describe and ultimately change the world.

That problem is one that applies even beyond the issue of embodiment. Embodiment makes specific claims about the nature of symbols, but those claims are not necessary to describe the problem with text-based theories of meaning. The primary issue is that words need to be connected to our concepts and then to the outside world in some way \citep{Harnad1990}. That is the critical link that must be included somewhere in every psychological theory of meaning. Note that we are not arguing that authors have been wrong in what they say about embodiment---there \emph{is} a conflict between embodiment and distributional semantics. We are merely pointing out that those issues apply more broadly, whether or not one thinks of concepts as embodied. 

\section{Conclusion} \label{sec_conclusion}

There is a long tradition in cognitive science of theorists claiming that such-and-such a computational paradigm cannot do such-and-such a task or reach a particular cognitive achievement. The record of such predictions is spotty. Often, the particular model criticized was replaced by future versions that were much better. By using a novel architecture, changing the learning algorithm, providing massive amounts of data, and so on, the putative impossible task turned out to be possible---at least to a reasonable degree of accuracy. We do not seek to join these ranks. Our point is not that text-based NLP models can't achieve interesting and important things; they surely have already, as NLP systems are becoming increasingly prominent in our daily lives (intelligent assistants, dialogue systems, machine translation, etc.). They will continue to advance and accomplish more important things. But they alone will not form the basis of a psychological theory of word meaning.

This may not concern researchers and practitioners seeking to optimize performance on particular tasks. We are not suggesting that NLP should switch its focus to building models of psychological semantics, at least not in every case. If one has large quantities of training data, it may be a very good idea to develop a task-specific model using standard approaches, or fine-tune a language model on that specific task. For example, if the goal is to develop a question answering model for a specific domain, and one has many thousands of question-answer pairs for training, large-scale pattern recognition may well be sufficient. Our arguments in this paper will have little relevance to such cases.

In other cases, a model of psychological semantics is a higher bar worth reaching for, with real payoff in terms of performance. We will not rehash the limitations of text-based NLP systems as psychological models. However, it is worth considering whether embracing a more psychologically motivated semantics would improve performance in future language applications. To understand language productively and flexibly, to produce reasonable responses to novel input, and to hold actual conversations will likely require something closer to a conceptually based compositional semantics of the sort that people have \citep{Marcus2019}. We make the following suggestions.

First, semantic representations need to be based on \emph{content}, information that makes contact with the world, and not just words connected to words. No matter how sophisticated the statistics or measure that links one word to others, word relations do not provide the basis for being able to talk about actual things and to get information from language. NLP models will need to move beyond pattern recognition and more firmly root themselves in concepts.

Second, word meanings have internal structure. You do not know what a dog is merely by knowing that it is connected to leashes, cats, mammal, leg, fur, toy, barking, etc. Your knowledge must be structured so that you know toys are things that dogs play with, fur is their body covering, mammal is a category they fall into, and so on. In identifying dogs, it is helpful to know that one of their parts is four legs. However, one must also understand in more detail what a dog's leg is and what it means to be a part. A dog's head next to four table legs does not add up to a dog, nor are those legs part of the dog. The relations between concepts and their constituents must be somehow encoded in order for the representation to work \citep{Brachman1979}.

This is accomplished by humans in part through a huge ``front end'' to their language learning, namely the perceptual-motor apparatus and knowledge of the world it provides. When a child first learns the word \emph{lion}, it is almost certainly while viewing a representation of or an actual lion. The child can perceive the parts, overall shape, color, sound, and possibly behaviors of the lion, without linguistic input. Indeed, most studies of child word learning use the method of \emph{ostension} to teach words: pointing at an object and labeling it. Children's sophistication in interpreting such experiences is impressive \citep[e.g.,][]{Markman1989}. The result is that they do not need verbal information in order to learn a great deal about what a lion is and how to identify one in the future. No one needs to describe the lion's face or say that the face is part of the lion, because that is directly learned via perception. Indeed, it is doubtful that any verbal description could adequately communicate what we know about lions' faces. Achieving such inferences with a hybrid visual-language model is an exciting possibility, albeit a difficult one to achieve. 

When parents in our culture do provide verbal instruction, it is often specifically labeled. For example, when teaching a word for a general concept, parents will often mention more specific examples and the set-superset relation that connects them, like ``Chairs, tables, and sofas are all kinds of furniture'' \citep{Callanan1990}. Such a statement indicates hierarchical relations between categories and suggests that chairs, tables, and sofas are at the same level (\emph{co-hyponyms}) under the umbrella category of furniture. They may also provide information such as ``Kitties say `meow','' which also provide the relation between \emph{kitten} and \emph{meow}, a relation that is different than the ones between \emph{kitten} and \emph{animal} or \emph{kitten} and \emph{fur}. Knowing the specific relations between objects and properties---taught via sentences including the words for those objects and properties---gives much more information about the world than simply knowing words' textual relations. This is not to claim that all knowledge is acquired through explicit instruction or propositional relations (though for those who attend school through college or post-graduate education, it is in fact the source of a tremendous amount of information) but rather to point out that this is one important way that world knowledge is transmitted, and theories of semantic knowledge need to accommodate it. If one's model has a misconception about unicorns, even after training on a corpus of 500 billion words, it would be awfully convenient to be able to just tell it, ``All unicorns have exactly one horn,'' as we can with people.

Contemporary multi-modal models, such as ones that learn to recognize objects and carry out instructions (see Figure \ref{fig_gSCAN}), are taking real steps toward some of these goals. It might well be that one can say that these models have a semantics. But as potential theories of psychological semantics, their linguistic abilities are usually too limited and too tied to specific patterns in their training experience. Looking towards the future, a robot that learns language while interacting with objects in the world (as well as receiving textual input) might well develop a semantics such that words successfully relate to things in the world, and the robot can describe the world with its language. Saying whether the robot's representations are functionally the same as human speakers' representations would require a detailed comparison of its abilities and the representations' internal structure. At present, these visual world models have simple linguistic representations that don't seem adequate as descriptions of human meanings, but it remains to be seen how these models develop.

\subsection{Final Conclusion}

Building a complete model of human word meaning requires, to some degree, building a conceptual structure of the world that people live in. Parts of that conceptual structure are linked to words, such that words pick out categories, properties, or relations. It doesn't seem likely that one can build such a structure out of textual statistics or word predictions, although sentences would certainly be one input to a learning system trying to build a coherent structure of the world. Such a system would have to try to interpret the sentences, identifying the relations, categories, and properties they describe. Most text-based systems do not do that or even try to do that. It is worth exploring how much text-based systems can do, because sampling even internet-scale data is easier than building the artificial intelligence to learn the detailed information in word meaning. Humans, with the advantage of perception, action, and reasoning are able to build complex knowledge structures, and these in turn form part of the basis of word meanings. Computational models of meaning will have to also form such structures if they are to be adequate psychological theories of meaning and, we propose, if they are to become sophisticated tools that can produce and understand language more broadly.

\section*{Authors' Note}
Order of authorship is alphabetical. We thank Marco Baroni, Gemma Boleda, Sam Bowman, Tammy Kwan, Gary Lupyan, Maxwell Nye, Josh Tenenbaum, and Tomer Ullman for helpful comments on an early draft. B. Lake's contribution was partially funded by NSF Award 1922658 NRT-HDR: FUTURE Foundations, Translation, and Responsibility for Data Science, and DARPA Award A000011479; PO: P000085618 for the Machine Common Sense program, both to NYU. A previous version of this manuscript appeared on arXiv.org (2008.01766).

\bibliographystyle{apacite}
\bibliography{greg,library_clean_lower}
\end{document}